\documentclass[english]{cg}               

\usepackage[utf8]{inputenc}

\usepackage{a4wide}
\usepackage{amsmath}
\usepackage{amssymb}
\usepackage{pslatex}
\usepackage{array}
\usepackage{multirow}
\usepackage{psboxit}
\usepackage{fancybox}
\usepackage{graphicx}
\usepackage{color}
\usepackage{float}

\usepackage{times}
\usepackage{epsfig}
\usepackage{graphicx}
\usepackage{amsmath}
\usepackage{amssymb}
\usepackage[small, rm]{subfigure}

\usepackage[pagebackref=true,breaklinks=true,letterpaper=true,colorlinks,bookmarks=false]{hyperref}

\def\etal{et~al.}

\begin{document}

\textheight=235mm
\voffset=-32mm
\hoffset=-8mm

\def \clip{\mathrel{\mathrm{clip}}}
\def \det{\mathop{\mathrm{det}}\nolimits}
\def \dim{\mathop{\mathrm{dim}}\nolimits}
\def \dist{\mathop{\mathrm dist}\nolimits}
\def \ease{\mathop{\mathrm{ease}}\nolimits}
\def \fac{\mathop{\mathtt{fac}}\nolimits}
\def \ggT{\mathop{\mathtt{ggT}}\nolimits}
\def \op{\mathrel{\mathrm{op}}}
\def \outcode{\mathop{\mathrm{outcode}}\nolimits}
\def \path{\bp C}
\def \pixel{\mathop{\mathrm{pixel}}\nolimits}
\def \profile{\bp S}
\def \rank{\mathop{\mathrm{rank}}\nolimits}
\def \round{\mathrel{\mathrm{round}}}
\def \sgn{\mathop{\mathrm{sgn}}\nolimits}
\def \sinc{\mathrel{\mathrm{sinc}}}
\def \spanof{\mathop{\mathrm{span}}\nolimits}
\def \spur{\mathop{\mathrm{spur}}\nolimits}
\def \stencil{\mathop{\mathrm{stencil}}\nolimits}
\def \supp{\mathop{\mathrm{supp}}\nolimits}
\def \weci{\mathop{\mathrm{wec}}\nolimits}
\def \wec{\mathop{\mathrm{WEC}}\nolimits}
\def \zbuff{\mathop{\mathrm{zBuffer}}\nolimits}

\def \go {\overline{\g }\, }
\def \gu {\underline{\g }\, }
\def \po {\overline{\varphi }\, }
\def \pu {\underline{\varphi }\, }
\def \bis {,\ldots ,}
\def \ul{\underline}
\def \ol{\overline}
\def \andop {\ensuremath{\wedge}}
\def \orop  {\ensuremath{\vee}}
\def \exorop  {\ensuremath{\widehat{\;}}}
\def \implop  {\ensuremath{\rightarrow}}
\def \equivop  {\ensuremath{\leftrightarrow}}
\newcommand{\negop}[1]{\ensuremath{\neg #1}}
\def \minusop {\backslash}
\def \mynl{~\\[-8mm]}
\def \e {\varepsilon }
\frenchspacing
\hyphenation {
}
%

%
%
\def \eop {{\hfill \vrule height7pt width7pt depth0pt \medskip }}
%
%
\def \strich {\vrule height7.5pt width0.2pt depth0pt}
\def \strichh {\vrule height6.5pt width0.2pt depth0pt}
\def \stricht {\vrule height7pt width0.6pt depth0pt}
%
%
\def \HH {{\mathbb H}}
\def \NN {{\mathbb N}}
\def \QQ {{\mathbb Q}}
\def \ZZ {{\mathbb Z}}
\def \RR {{\mathbb R}}
\def \BB {{\mathbb B}}
\def \CC {{\mathbb C}}
%
%
\renewcommand{\Re}{\mathfrak{Re}}
\renewcommand{\Im}{\mathfrak{Im}}
%
%
\def \l {\lambda }
\def \L {\Lambda }
\def \m {\mu }
\def \a {\alpha }
\def \A {\it A }
\def \b {\beta }
\def \e {\varepsilon }
\def \d {\delta }
\def \D {\Delta }
\def \g {\gamma }
\def \k {\kappa }
\def \m {\mu}
\def \p {\phi }
\def \pa {\partial}
\def \P {\Phi }
\def \th {\theta }
\def \Th {\Theta }
\def \r {\varrho }
\def \Om {\Omega}
\def \om {\omega}
\def \si {\sigma}
\def \Si {\Sigma}
\def \ka {\kappa}
\def \lra {\longrightarrow}
\def \lla {\longleftarrow}
\def \ra  {\rightarrow}
\def \la  {\leftarrow}
\def \Lra {\Longrightarrow}
\def \Lla {\Longleftarrow}
\def \Ra  {\Rightarrow}
\def \La  {\Leftarrow}
\def \leria  {\leftrightarrow}
\def \Leria  {\Leftrightarrow}
\def \da    {\downarrow}
\def \ua    {\uparrow}
%
%
\def \cA {{\cal A}}
\def \cB {{\cal B}}
\def \cC {{\cal C}}
\def \cD {{\cal D}}
\def \cE {{\cal E}}
\def \cF {{\cal F}}
\def \cG {{\cal G}}
\def \cH {{\cal H}}
\def \cI {{\cal I}}
\def \cJ {{\cal J}}
\def \cK {{\cal K}}
\def \cL {{\cal L}}
\def \cM {{\cal M}}
\def \cN {{\cal N}}
\def \cO {{\cal O}}
\def \cP {{\cal P}}
\def \cQ {{\cal Q}}
\def \cR {{\cal R}}
\def \cS {{\cal S}}
\def \cT {{\cal T}}
\def \cU {{\cal U}}
\def \cV {{\cal V}}
\def \cW {{\cal W}}
\def \cX {{\cal X}}
\def \cY {{\cal Y}}
\def \cZ {{\cal Z}}
%
%
%
\def\dps{\displaystyle}
\def\txs{\textstyle}

%
%
\def \Rom#1 {\uppercase\expandafter{\romannumeral #1}}
\def \rom#1 {\expandafter{\romannumeral #1}}
\newcommand{\gaussB}[1]{\left\lfloor #1 \right\rfloor}
\newcommand{\gaussT}[1]{\left\lceil #1 \right\rceil}
%
%
\def \({\left( }
\def \){\right) }
%
%
\newcommand{\bp}[1]{{\mathbf #1}}
\newcommand{\pv}[1]{{\mathbf #1}}
\newcommand{\bv}[1]{{\vec{\mathbf #1}}}
\newcommand{\nv}[1]{{\hat{\mathbf #1}}}
\newcommand{\sv}[1]{{\vec{#1}}}
\newcommand{\Vecc}[2]{{\renewcommand{\arraystretch}{0.8}%
    \begin{pmatrix}#1\\#2\end{pmatrix}}}
\newcommand{\Veccc}[3]{{\renewcommand{\arraystretch}{0.8}%
    \begin{pmatrix}#1\\#2\\#3\end{pmatrix}}}
\newcommand{\Vecccc}[4]{{\renewcommand{\arraystretch}{0.8}%
    \begin{pmatrix}#1\\#2\\#3\\#4\end{pmatrix}}}
\newcommand{\HVec}[2]{{\renewcommand{\arraystretch}{0.8}%
    \begin{bmatrix}#1\\#2\end{bmatrix}}}
\newcommand{\HVecc}[3]{{\renewcommand{\arraystretch}{0.8}%
    \begin{bmatrix}#1\\#2\\#3\end{bmatrix}}}
\newcommand{\HVeccc}[4]{{\renewcommand{\arraystretch}{0.8}%
    \begin{bmatrix}#1\\#2\\#3\\#4\end{bmatrix}}}
\newcommand{\hVecc}[2]{{\left( #1, #2\right)}}
\newcommand{\hVeccc}[3]{{\left( #1, #2, #3\right)}}
\newcommand{\hVecccc}[4]{{\left( #1, #2, #3, #4\right)}}
\newcommand{\edge}[2]{\ol{\bp #1\bp #2}}
\newenvironment{mcases}{\left\{\begin{matrix}}{\end{matrix}\right\}}
\newenvironment{rcases}{\left.\begin{matrix}}{\end{matrix}\right\}}
%
%
\newcommand{\pd}[2]{\frac{d #1}{d #2} }
\newcommand{\PD}[2]{\frac{\partial #1}{\partial #2} }
\newcommand{\PDDq}[2]{\frac{\partial^2 #1}{\partial #2^2} }
\newcommand{\PDD}[3]{\frac{\partial^2 #1}{\partial #2 \partial #3} }
\newcommand{\PDDDq}[2]{\frac{\partial^3 #1}{\partial #2^3} }
\newcommand{\DQ}[2]{D_{#2}[#1]}
\newcommand{\sfrac}[2]{\mbox{$\frac{#1}{#2}$}}
\newcommand{\lfrac}[2]{\frac{\dps #1}{\dps #2}}
\newcommand{\quat}[1]{{\underline{\mathbf #1}}}
%
%
\newcommand{\Betr}[1]{ \left| #1 \right| }
\newcommand{\Norm}[1]{ \left\| #1 \right\| }
\newcommand{\Normf}[2]{ \left| #1 \right|_{#2} }
\newcommand{\Vprod}[2]{{#1\times#2}}
\newcommand{\IProd}[2]{{\(#1\cdot#2\)}}
\newcommand{\BigIProd}[2]{ \left\langle \; {#1} \mid {#2} \; \right\rangle }
\newcommand{\brdfl}[3]{f_r(#1, #2 \ra #3, \lambda)}
\newcommand{\brdf}[3]{f_r(#1, #2 \ra #3)}
\newcommand{\brdfs}[2]{f_r(#1 \ra #2)}
\newcommand{\fff} {\text{\sf I}}
\newcommand{\sff} {\text{\sf II}}
\newcommand{\tangent}[1]{\nv t_{\bp #1}}
\newcommand{\cnormal}[1]{\nv n_{\bp #1}}
\newcommand{\binormal}[1]{\nv b_{\bp #1}}
\newcommand{\curvature}[1]{\k_{\bp #1}}
\newcommand{\tanplane}[2] {{\bp T}_{#1}(#2)}
\newcommand{\Ktensor}{{\cal K}}
\newcommand{\cov}{\nabla}
\newcommand{\hess}{\text{\sf Hess\:}}
\newcommand{\tr}{\text{\sf tr}\:}
\newcommand{\ft}[1]{\mathop{\mathbf FT}\nolimits[#1]}
\newcommand{\ftinv}[1]{\mathop{{\mathbf FT}^{-1}}\nolimits[#1]}
\newcommand{\dft}[1]{\mathop{\mathbf FT}\nolimits[#1]}
\newcommand{\dftinv}[1]{\mathop{{{\mathbf FT}}^{-1}}\nolimits[#1]}
\newcommand{\wong}[2]{#1{\mathbf d}#2{\mathbf v}}
\newcommand{\brodlie}[2]{{\it E}_{#1}^{#2}}
\newcommand{\Eig}{\mathop{\mathbf Eig}\nolimits}
\newcommand{\diag}[1]{\mathop{\mathbf diag}\nolimits[#1]}

\def \grad      {\text{\sf grad}\:}
\def \laplace   {\nabla}
\def \diver     {\text{\sf div}\:}
\def \Mkr       {\text{\sf H}}               
\def \Gkr       {\text{\sf K}}               
\def \dil       {\oplus}
\def \ero       {\circleddash}
\def \id        {\text{Id}}

%
%
\def \ine {\mathop{\stackrel{\circ}{e}}\nolimits}
\def \vls {\rule[-6mm]{.2pt}{8mm}}
\def \vlf {\rule[-6mm]{.2pt}{10mm}}
\def \vlt {\rule[-4mm]{.2pt}{6mm}}
\def \iff {\Leftrightarrow}
\def \id  {\text{\sf Id}}
\def \relates {\hat{=}}
\renewcommand{\bar}{\overline}
\newcommand{\boxpar}[1]{\centerline{\fbox{\parbox{.8\textwidth}{#1}}}}
\newcommand{\boxline}[1]{\centerline{\fbox{#1}}}
\newcommand{\boxeq}[1]{{\centerline{\fbox{$\displaystyle #1$}}}}
%
%
\newcommand{\tb}[1]{{\tt{\mathbf #1}}}
%
%
\def \bezier    {B\'ezier}
\def \sequin    {S\'equin}
\def \poly      {{\cal P}}
\def \opengl    {{OpenGL}}
\def \oiv       {{Open\-Inventor}}

\newcommand{\nl}{\newline}
\newcommand{\hf}{\hfill}
\newcommand{\hr}{\smallskip\hrule\smallskip}

\dbDocumentType{Technical Report}
\dbTitle{Hand Tracking based on Hierarchical Clustering of Range Data}
\dbAuthor{Roberto Cespi, Andreas Kolb, Marvin Lindner}
\dbVersion{\today}
\dbDisclosure{0}

\cgDeckblatt

\subsection*{Abstract}
\begin{quote}
  Fast and robust hand segmentation and tracking is an essential basis for
  gesture recognition and thus an important component for contact-less
  human-computer interaction (HCI). Hand gesture recognition based on 2D video
  data has been intensively investigated. However, in practical scenarios
  purely intensity based approaches suffer from uncontrollable environmental
  conditions like cluttered background colors.

  In this paper we present a real-time hand segmentation and tracking
  algorithm using Time-of-Flight (ToF) range cameras and intensity data.  The
  intensity and range information is fused into one pixel value, representing
  its combined intensity-depth homogeneity. The scene is hierarchically
  clustered using a GPU based parallel merging algorithm, allowing a robust
  identification of both hands even for inhomogeneous backgrounds. After the
  detection, both hands are tracked on the CPU. Our tracking algorithm can
  cope with the situation that one hand is temporarily covered by the other
  hand.
\end{quote}

\section{Introduction}

Gesture-based real-time human-computer interaction requires a fast and robust
segmentation of the human hands~\cite{mitra07gesturesurvey}. Classical
approaches are based on 2D intensity or color images. However, this kind of
techniques suffers from low efficiency and the lack of robustness in case of
cluttered scenes or if applied under varying lighting conditions. Addressing
practical application scenarios, techniques capable of handling both effects
are strongly required; frequently applied simplifications, e.g. restricted
lighting or material conditions~\cite{keskin05gesture}, or marker- or
glove-based approaches~\cite{starner98signlanguage} are hardly applicable.

One major approach to overcome the problems of segmenting intensity or color
image sequences for gesture recognition purposes is to use additional depth
information, delivered by laser range systems~\cite{heisele99range}, stereo
cameras~\cite{lee10gameinterface} or structured light range acquisition
systems~\cite{malassiotis01gesture}. The major drawback of all these
approaches is the comparably expensive sensing hardware and the significant
space requirement, which is due to systematic constraints, e.g. the baseline
required for stereo techniques or structured light, or mechanical setups in case of laser
range scanners.

Recently, the {\em Time-of-Flight (ToF)} technology, based on measuring the
time that light emitted by an illumination unit requires to travel to an
object and back to a detector, has been manufactured as highly integrated
\emph{ToF cameras}.  Unlike the other 3D acquisition systems, ToF cameras are
very compact. ToF-cameras are realized in standard CMOS or CCD technology and
thus can be cost efficiently manufactured~\cite{kolb10tof, xu99smartpixel}.
ToF-cameras have been successfully applied in the context of man-machine
interaction, e.g. for facial tracking~\cite{haker07facial}, for touch-free
navigation in 3D medical applications~\cite{soutschek08gesture}, upper-body-gesture~\cite{Holte08:Gesture} and hand-gesture
recognition~\cite{ghobadi08robot}.

In this paper, we introduce a hand segmentation approach based on a
hierarchical clustering technique. Using hierarchical clustering is beneficial,
since the final number of clusters of the scene delivering the ``best'' hand
segmentations highly depends on the scene complexity and thus can not be
determined beforehand. To achieve a high performance, we adopt the GPU-based
clustering approach introduced by Chiosa and Kolb~\cite{chiosa11multilevel} to
cluster fused range-intensity images. In this context, we introduce a novel
homogeneity criterion for range-intensity images. Our hand segmentation and
tracking system is capable of robustly detecting and tracking both hands, even
under the condition one hand is temporarily covered by the other hand or in case a distructing object, e.g. a third hand appears.

The reminder of the paper is structured as follows. In Sec.~\ref{s:priorwork}
we discuss major prior work on segmentation and tracking
techniques. Sec.~\ref{s:overview} gives an overview on our hand tracking
system, followed by detailed discussions on the applied hierarchical
clustering method (Sec.\ref{s:clustering}) and the hand tracking
(Sec.~\ref{s:tracking}). Sec.~\ref{s:results} presents some experimental
results of our tracking approach and in Sec.~\ref{s:conclusion} we draw some
final conclusions.


\section{Related Work}
\label{s:priorwork}

\subsection{Segmentation}
\label{s:priorwork.segmentation}
The most important basis for a hand tracking system is a robust and fast
clustering algorithm in order to segment the hands from their environment. The
Graph Cut algorithms, or Graph-based clustering algorithms \cite{Boy01:GC},
used by Schoenberg~\etal~\cite{Schoen10SU}, delivers good clustering
results. However, with a runtime of approximately 3-4 FPS (frames per second) for
a $204^2$ pixel sized image, this method is unsuitable for real-time
applications. Another, simpler approach is used by
Breuer~\etal~\cite{Breuer07:Gesture}, where only the depth information is used
to segment one hand from the background. By depth keying the background, only
the nearest 3D point cloud will be detected as hand. This method however is too inflexible because the hand is restricted to a certain spatial
position. Furthermore a hand segmentation based on range data only is too
error-prone. Holte~\etal~\cite{Holte08:Gesture} present a motion detection
approach, where only the moving arms are segmented. In order to extract the moving
arms Holte~\etal used 3D double difference images and represent this data by their 
shape context scheme. Tsap~\cite{Tsap02:GT} and Tsap and
Shin~\cite{Tsap04:3DHT} use a connected component analysis of skin-colored
pixels, to segment the hands and the face. Only clusters with a certain shape
will be selected as possible hand clusters. In \cite{GLHL07:HS} Ghobadi~\etal
propose a segmentation technique which combines two clustering approaches to
cluster fused range intensity images: k-means and expectation
maximization. The drawback of this method is the necessity to decide on the
number of clusters beforehand.

\subsection{Tracking}
\label{s:priorwork.tracking}
Tsap and Shin \cite{Tsap04:3DHT} use a simple depth analysis to distinguish
the hand from the face, defining the skin colored cluster nearest to the
camera as the hand. Furthermore they limit the depth space search for the
hand dynamically by using the motion history of the last frames. In
\cite{Tsap02:GT}, Tsap tracks only non-static skin-colored objects by applying
a set of filters on color, motion, and range data. After segmentation,
Breuer~\etal~\cite{Breuer07:Gesture} apply a two-stage approach to detect the
hand based on principal component analysis and a refinement based on a model-fitting in object space.
Holte~\etal~\cite{Holte08:Gesture} use the shape context (see
Sec.~\ref{s:priorwork.segmentation}) of the segmented moving arms for gesture
recognition. A gesture is recognized by matching the current harmonic shape
context with a known set, one for each possible
gesture. Haker~\etal~\cite{haker07facial} use a simple classifier to detect
the human nose in fused intensity-range images for facial
tracking. Soutschek~\etal~\cite{soutschek08gesture} use an approach to detect
the hand as the closest object with regard to the ToF-camera. They combine thresholding
applied to depth and amplitude information and an explicit palm detection in order to
remove the forearm geometry. In \cite{ghobadi08robot}, Ghobadi~\etal apply
their k-means clustering technique proposed in \cite{GLHL07:HS} to
interactively track the hand on a frame-to-frame basis using haar-like
features in combination with AdaBoost.


\section{System Overview}
\label{s:overview}

\begin{figure}[t!]
  \begin{center}
    \includegraphics[width=.99\linewidth]{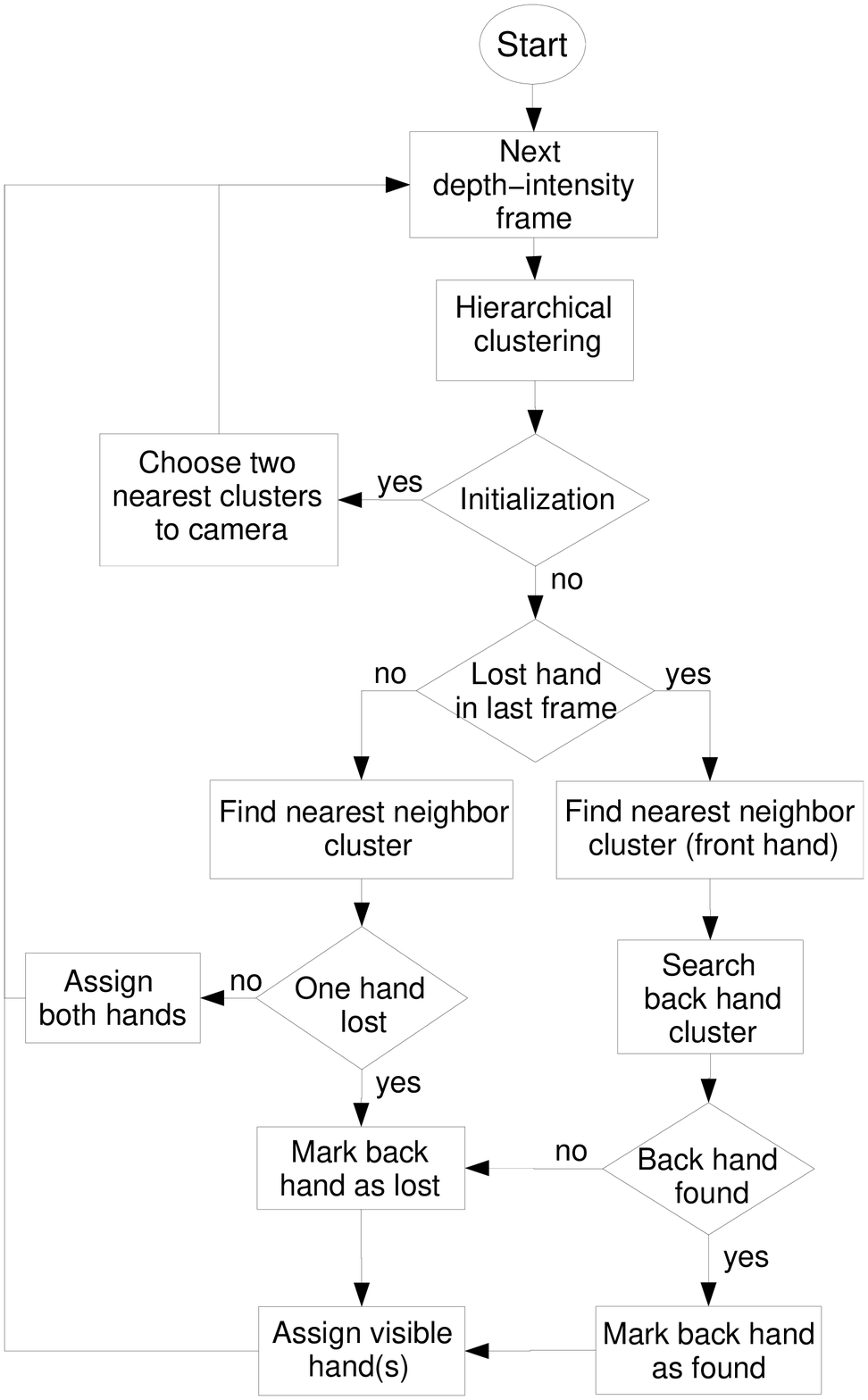} 
    \caption{Overview of the clustering and hand tracking algorithm.}
    \label{fig:fc}
  \end{center}
\end{figure}

The clustering and hand tracking algorithm is shown in
Fig.~\ref{fig:fc}. Summarizing the procedure, our system consists of the two
major components, the {\itshape hierarchical clustering} and the {\itshape
  tracking} component. 

\paragraph{Hierarchical Clustering } After receiving the range and intensity
information from the camera, the GPU based parallel merging algorithm clusters
the fused range-intensity data into regions (see
Sec.~\ref{s:clustering:pm}). The algorithm searches the mutual, optimal merge
partner for every region according to some homogeneity criterion. If this is
the case, both regions will be merged into one and the region size and data
will be updated. The clustering stops, when there are no optimal merge
partners left (see Sec.~\ref{s:clustering:mc}).

\paragraph{Tracking } The tracking algorithm analyzes the clustering
result. If the system is still in the initialization phase, the two clusters
nearest to the camera, which have a given minimum size, will be assigned as the 
first hand and the second hand (see Sec.~\ref{s:tracking:init}). After the
initialization, from the current set of clusters two new hand clusters are
chosen by comparing their homogeneity values to the hand-clusters from the
last frame. The hands are assigned trough a nearest neighbor method (see
Sec.~\ref{s:tracking:track}). If only one cluster was found, we assume that
one hand is covered by the other, the lost hand will be marked. The lost hand
will be searched in the subsequent frames and will only be marked as found, if
a cluster satisfies specific homogeneity and space criteria (see
Sec.~\ref{s:tracking:special}).

\section{Segmentation (Clustering)}
\label{s:clustering}

A hierarchical clustering algorithm, implemented on the GPU as a {\itshape parallel merging algorithm},
is used in order to segment the hands from their environment. The advantage of a hierarchical approach is based on the fact, that the
number of clusters, and thus the size and shape of the hand clusters, results
only from the applied homogeneity criterion. This means that no predefined number is necessary.

The parallel clustering method (see Sec.~\ref{s:clustering:pm}), as
well as the merging criterion (see Sec.~\ref{s:clustering:mc}) which is used to merge
pixels and clusters into new clusters, are discussed in the following
subsections.

\subsection{Parallel Merging Algorithm}
\label{s:clustering:pm}
The Parallel Merging Algorithm is a hierarchical clustering method
\cite{WR11:RG}, where initially each pixel represents a region (cluster) with
his own region-ID and region characteristics.

For merging two regions $\cR_1, \cR_2$, they have to satisfy a given boolean criterion based on a homogeneity descriptor $\bv w^i=\hVeccc{w^i_1}{\ldots}{w^i_N},\;i=1,2$ consisting of $N$ components, i.e.
\begin{equation}
\label{eq:mergecrit}
\Betr{w^1_i-w^2_i} \leq t_i,\; i=1\bis N.
\end{equation}
Based on the the individual homogeneity components $w_i$, the homogeneity difference between two regions is deduced
\begin{equation}
\label{eq:mergecrit2}
f(\cR_1,\cR_2) = \sum_{i=1}^N \alpha_i\cdot\Betr{w_{i}^1-w_{i}^2},
\end{equation}
where $\alpha_i$ is the weight of the individual homogeneity components. 

\begin{figure}[t!]
  \begin{center}
    \begin{tabular}{ccc}
      \subfigure[]{
        \includegraphics[width=.25\linewidth]{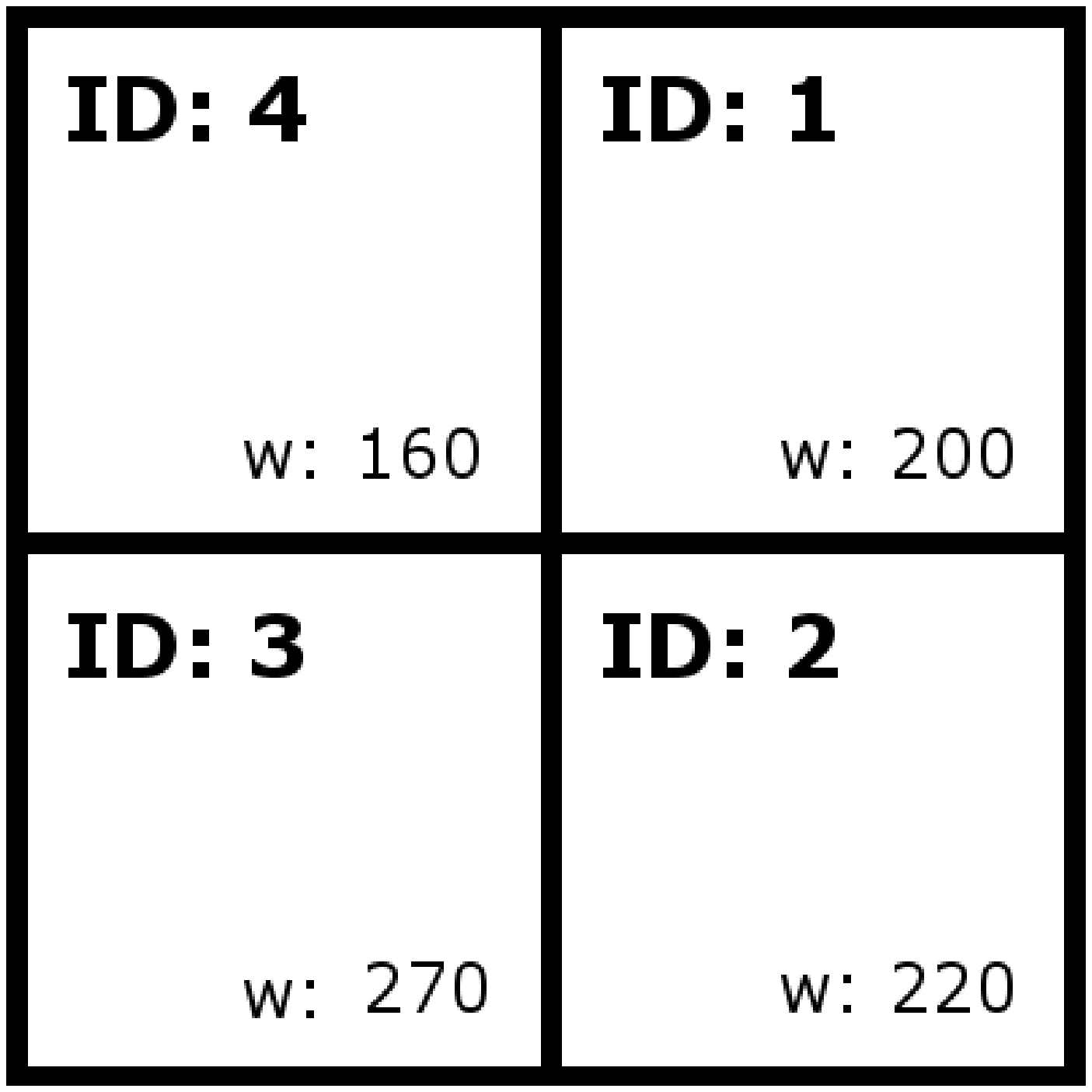}
        \label{fig:sm.1}
      }&
      \subfigure[]{
        \includegraphics[width=.25\linewidth]{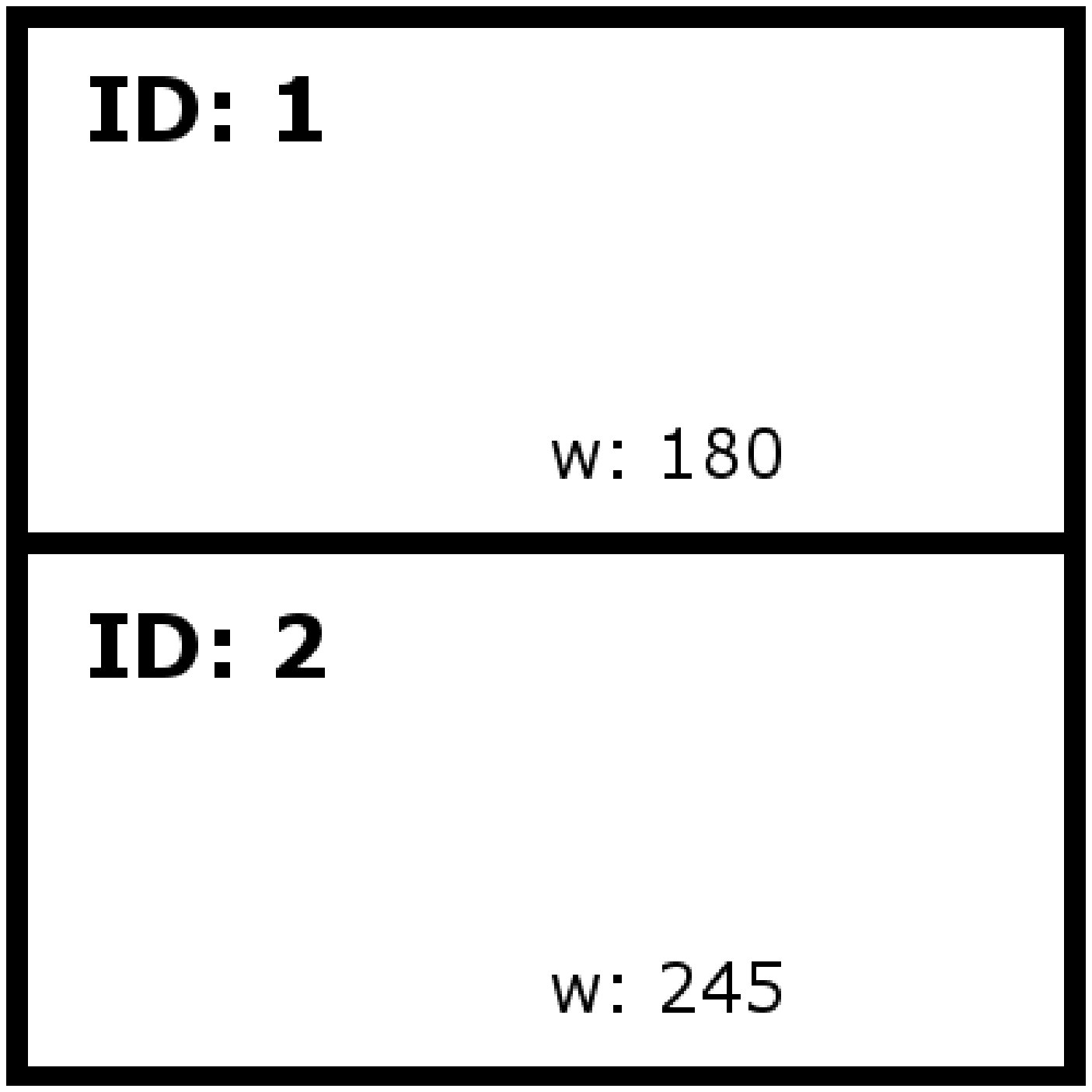}
        \label{fig:sm.2}
      }&
      \subfigure[]{
        \includegraphics[width=.25\linewidth]{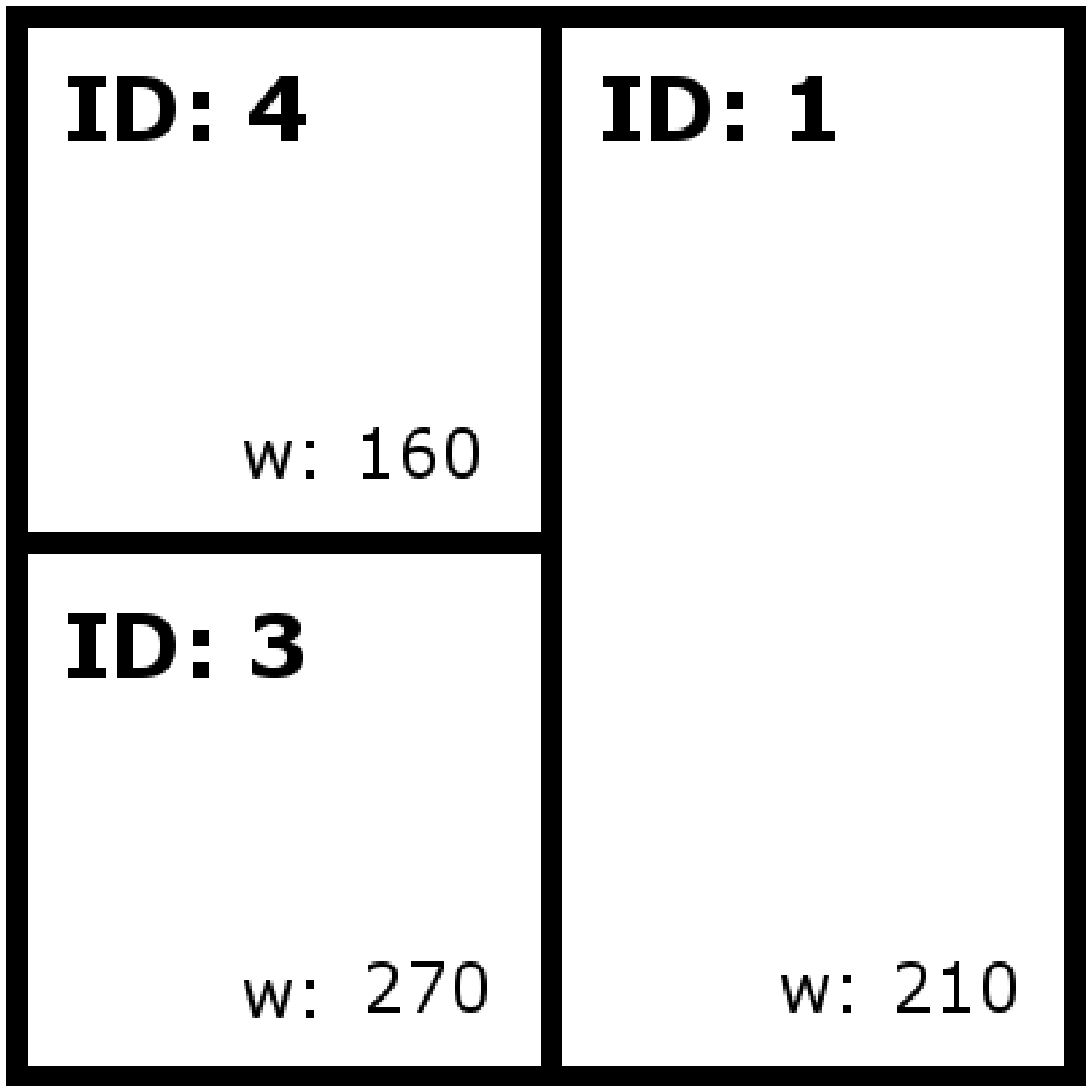}
        \label{fig:sm.3}
      }
    \end{tabular}

  \end{center}
  \caption{(a) A clustering into four regions, with their ID and
    homogeneity measure $w$. Threshold $t = 51$. (b) Merging result
    for sequential merging, starting merging from top left to bottom
    right. (c) Merging result for sequential merging, starting merging from
    top right to bottom left.}
  \label{fig:sm}
\end{figure}

Given a specific cluster $\cR$, all neighboring clusters $\cS$
that satisfy the homogeneity criterion are identified and the optimal merge
partner $\cR^\ast$ is selected, i.e.
$$
\cR^\ast=\arg\min_{\cS}\{f(\cR,\cS)\}.
$$
After identifying the optimal merge partner $\cR^\ast$ for $\cR$, both regions
will be merged into a single region and the homogeneity measure for the merged region is
determined. The new merged region gets the greater region ID of both merge
partners and its characteristics will be updated. However, with a sequential
merging algorithm, the merging result depends on the merging order
(see Fig.~\ref{fig:sm}). A better solution is therefore to check for all regions its optimal merge partner.
Two regions are merged only if the
optimal merge partner choice is mutual, i.e. if $(\cR^\ast)^\ast = \cR$.

\noindent
The merging of regions is performed in parallel and thus the following rules must be obeyed in order
to obtain a correct result \cite{WR11:RG}:
\begin{enumerate}
\item Each region can only merge with one other region at a time; that being
  the neighbor which best satisfies the merging criteria.
\item In case of an equal measure the neighbor with the larger ID is selected in
  order to prevent cyclic dependencies.
\item A merge choice must be mutual in order for two regions to merge.
\end{enumerate}  

\begin{figure}[t]
  \begin{center}
    \begin{tabular}{ccc}
      \subfigure[]{
        \includegraphics[width=.25\linewidth]{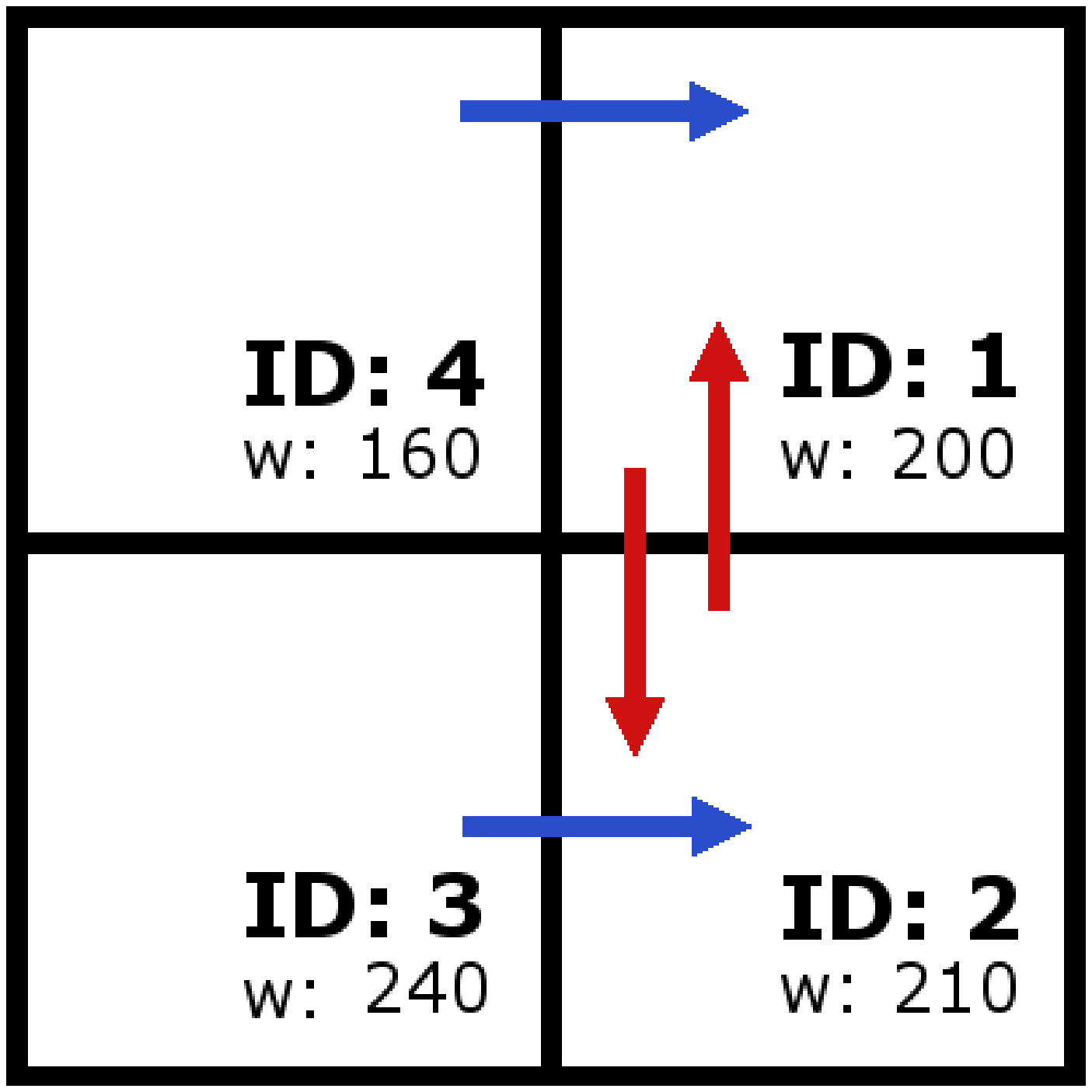}
        \label{fig:pm.1}
      }&
      \subfigure[]{
        \includegraphics[width=.25\linewidth]{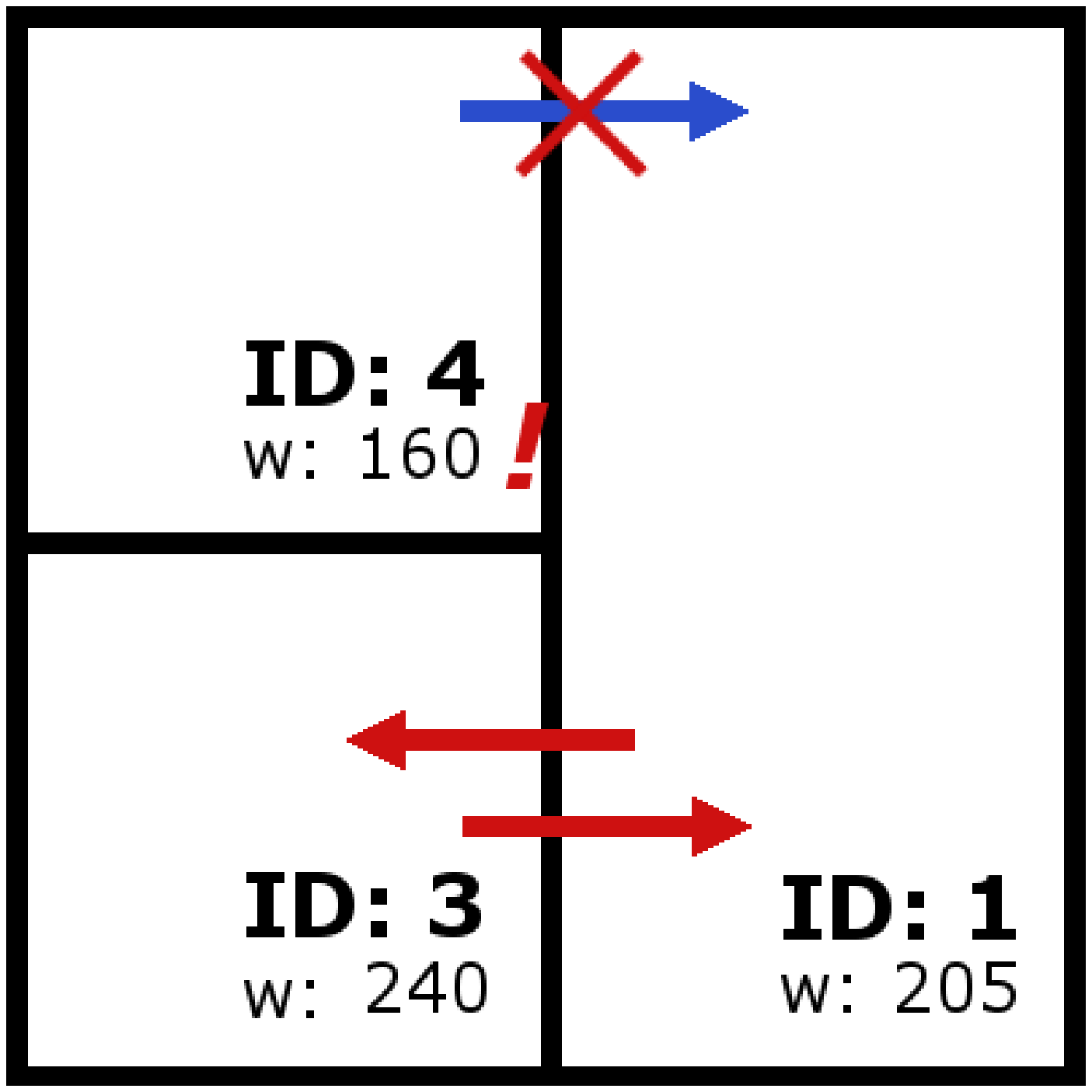}
        \label{fig:pm.2}
      }&
      \subfigure[]{
        \includegraphics[width=.25\linewidth]{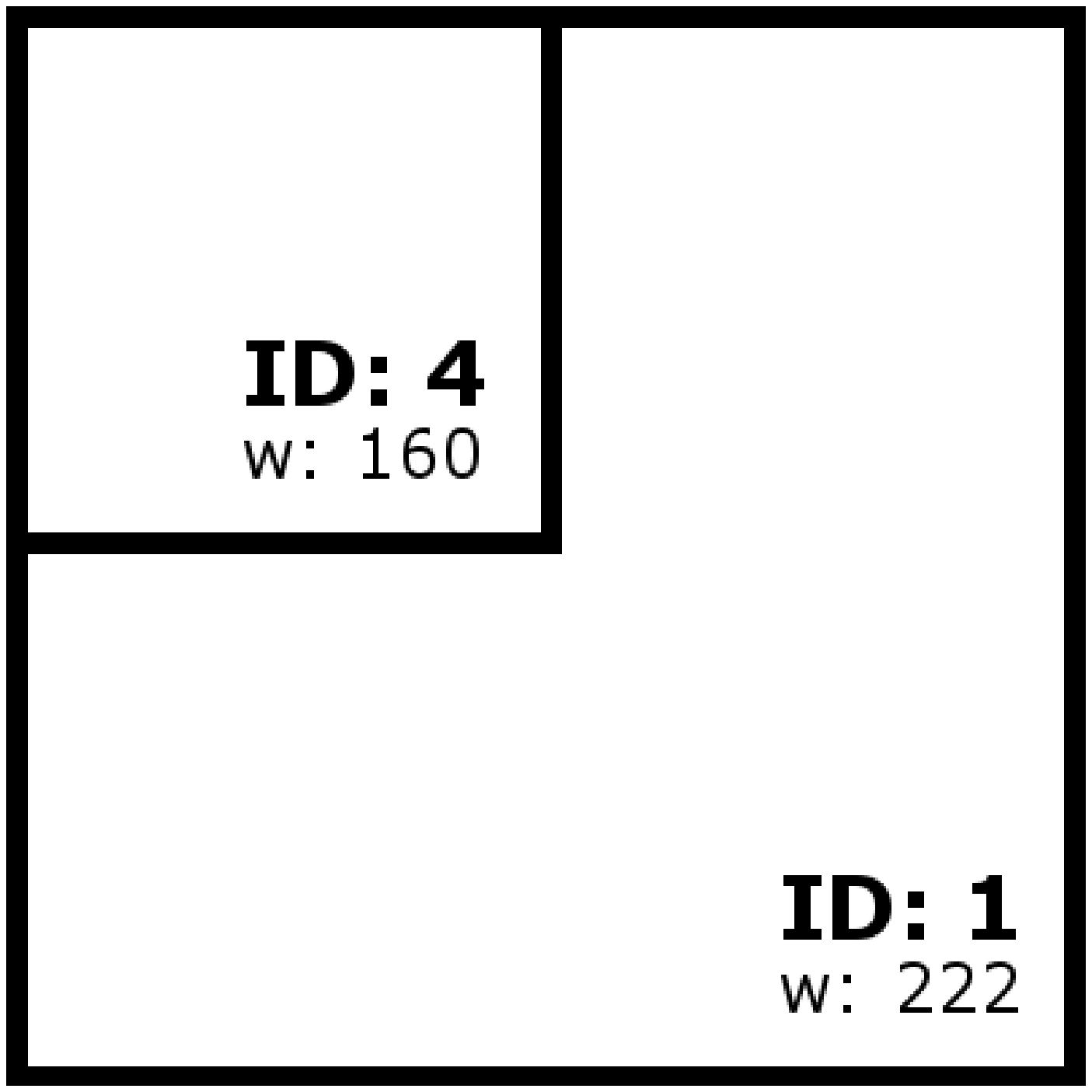}
        \label{fig:pm.3}
      }
    \end{tabular}

  \end{center}
  \caption{Threshold $t = 40$. (a) Mutual merging decision between region 1
    and 2. Region 3 and 4 have to wait until the next iteration. (b) Merging
    result from (a). Mutual merging decision between region 3 and 1. Region 4
    however does not satisfy the merging criteria anymore. (c) Final merging
    result.}
  \label{fig:pm}
\end{figure}

These rules restrict a region to only merge with a single neighbor during any 
merging iteration. The resulting region could otherwise be in violation
of the homogeneity requirements (see Fig.~\ref{fig:pm}).  By giving the new
region the greater ID (in combination with rule number 2), possible deadlocks can
be avoided. After merging two regions into a single region, its characteristics
will be updated.

A region that was unable to merge during a given iteration because its
selection was not mutual, may succeed in a subsequent iteration (see
Fig.~\ref{fig:pm}).

This iteration will be repeated until there is no optimal merge partner for
any region left, i.e. until all pairs of neighbors violate the merge criterion
(see Eq.~\eqref{eq:mergecrit}).

\subsection{Merging Criterion}
\label{s:clustering:mc}
The merging criterion, which is used to check the homogeneity between
neighboring regions is based on fusing intensity and range information into a
single value, similar to Ghobadi~\etal~\cite{GLHL07:HS}. The choosen
homogeneity descriptor $\bv w$ (s. Sec.~\ref{s:clustering:pm}) which characterizes a region consists of two components:
\begin{equation}
  \bv w_\cR = (z_\cR,\p_\cR),
\end{equation} 
where $z_\cR$ denotes the average $Z$-distance of the region with regard to the camera's optical axis and $\p_\cR$ the fused intensity and range value.

The characterizing entities for each region $\cR$, i.e. initially a single pixel, are
given by the distance $d_\cR$ in spherical sensor coordinates and the
reflected active light $I_\cR$ of a region/pixel. 

As the desired $\p_\cR$ needs to represent homogeneous
characteristics of a pixel region in a robust way, we apply the well known
dependency rule between intensity and the distance to derive a robust measure:
\begin{equation}
  I_\cR \propto \frac{1}{d_\cR^2} \Rightarrow 	\frac{1}{\sqrt{I_\cR}} \propto
  d_\cR \Rightarrow d_\cR \sqrt{I_\cR} \propto \text{const}.
\end{equation} 
Thus, $d_\cR \sqrt{I_\cR}$ is a nearly constant measure for every individual
surface and interpreting each $(d_\cR,I_\cR)$-pair as point in the Euclidean
plain, we derive $\p_\cR$ as follows:
\begin{equation}
  \label{eq:phi}
  \p_\cR = \arctan(d_\cR\sqrt{I_\cR}).
\end{equation}

When merging regions, the mean average of both regions $z_\cR$ and $\p_\cR$ will be used for the new region's homogeneity descriptor.
The values of the thresholds ($t_z$ and $t_\p$) for the merging criterion, as well the weights ($\alpha_z$ and $\alpha_\p$) needed for the homogeneity difference, are listed in Table \ref{tab:ths}.

The hierarchical merging is applied to each frame delivered by the
ToF-camera. The algorithm is realized using a GPU-implementation similar to
Chiosa and Kolb~\cite{chiosa11multilevel}, which has be introduced for mesh
and data clustering. We simply use a standard 4-neighborhood-size to define pixel
neighbors and integrate our homogeneity measure as the merge criterion.


\section{Hand Tracking}
\label{s:tracking}

The hand tracking algorithm has been implemented on the CPU and is divided
into two steps. Step one is the initialization, where the hands will be
identified. In step two, both hands will be tracked.

\subsection{Initialization}
\label{s:tracking:init}
\begin{figure}[b!]
  \begin{center}
    \subfigure[]{
      \includegraphics[width=.6\linewidth]{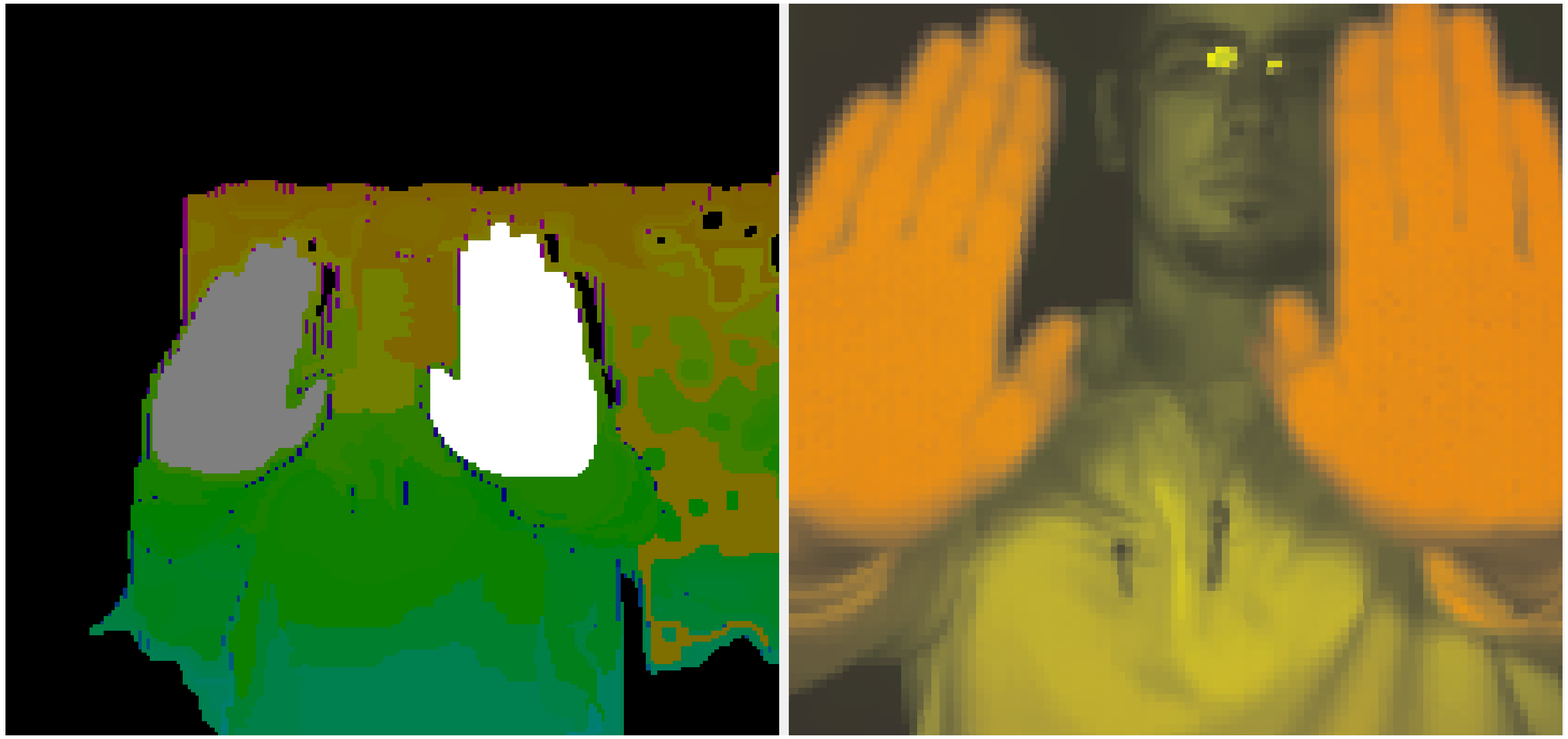}
      \label{fig:detec.1}
    }
    \subfigure[]{
      \includegraphics[width=.6\linewidth]{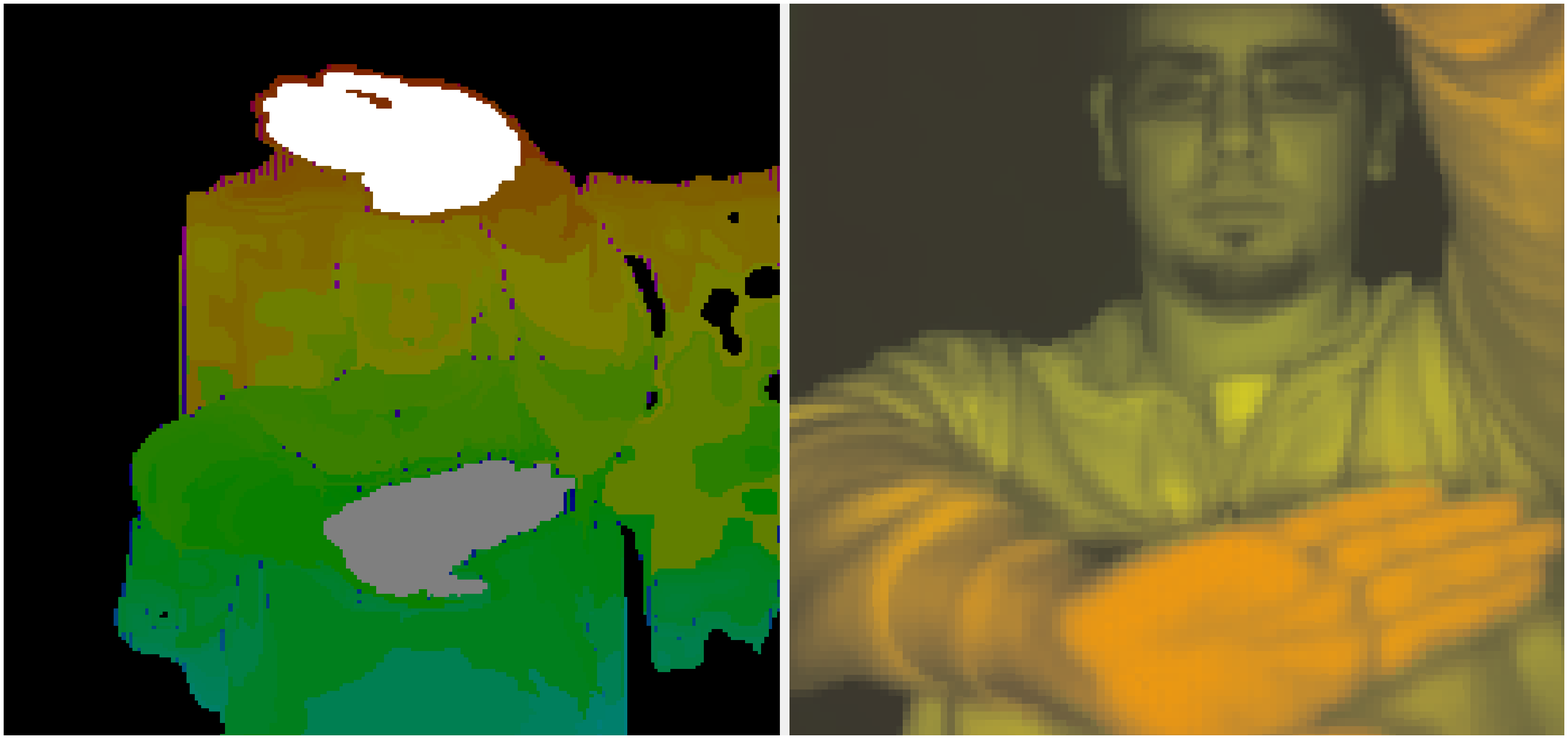}
      \label{fig:detec.2}
    }
    \subfigure[]{
      \includegraphics[width=.6\linewidth]{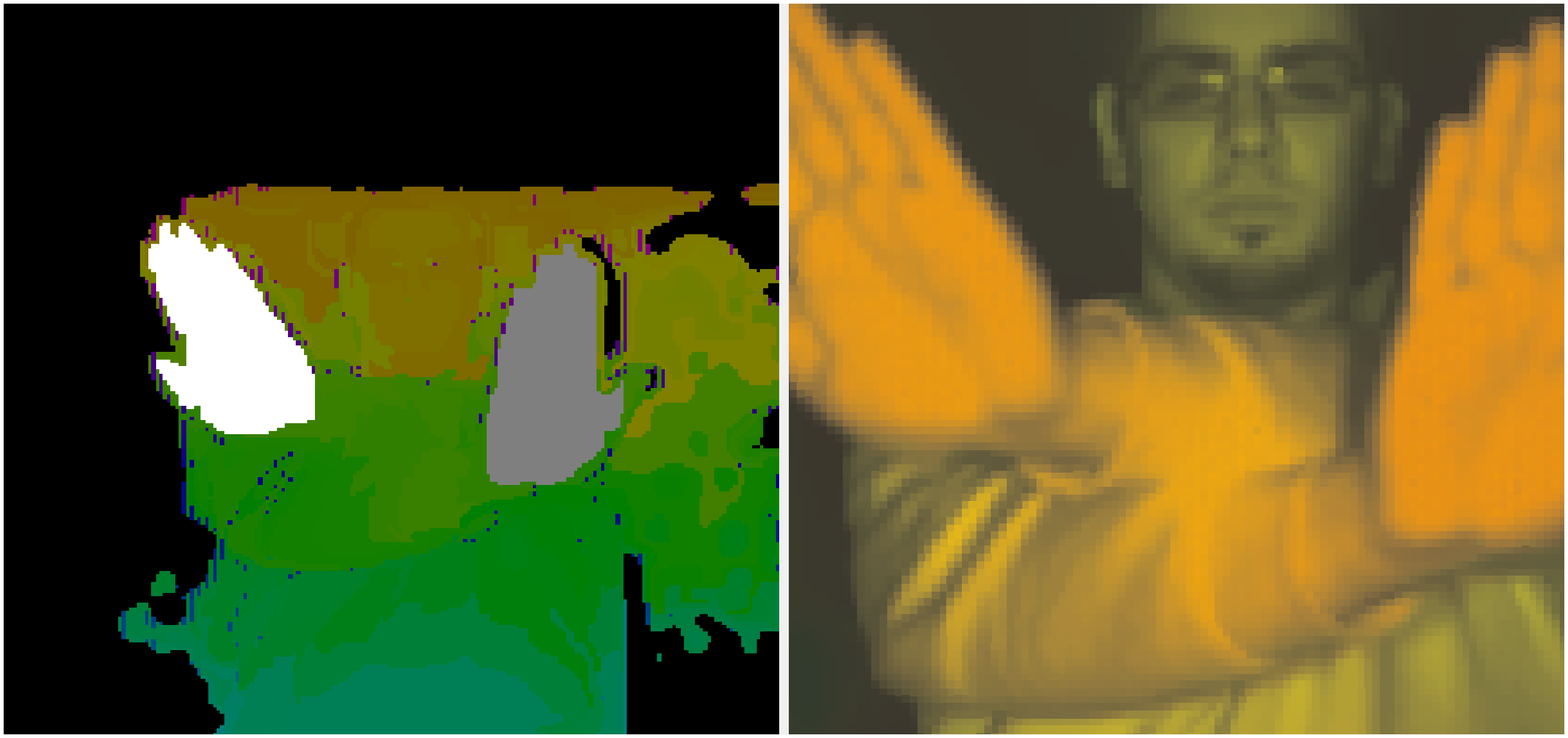}
      \label{fig:detec.3}
    }
    \caption[Hand-detection]{As long as both hands satisfy the criteria from
      the initialization step, they can always be assigned correctly. The
      white and the gray cluster resembles the first hand respectively the second
      hand (left: clustered image; right: intensity image).}
    \label{fig:detec}
  \end{center}
\end{figure}

In step one, the hands will be detected from the set of regions, that has been
clustered until convergence, i.e. until no further pair of neighboring clusters
fulfills the merge criterion (see Eq.~\eqref{eq:mergecrit}).  The two clusters which
will be identified as hands are determined by the following simple rules:
\begin{itemize}
\item closest to the ToF-camera and
\item the cluster size is above a given threshold.
\end{itemize}
Both regions will be then assigned as first hand (cluster $\cR_1$) and second
hand (cluster $\cR_2$). As long as these regions satisfy the mentioned
criteria, both hands can be clearly assigned in the successive image through a
nearest neighbor assignment (see Fig. \ref{fig:detec}).

After 30 frames, the initialization stops and the tracking is
initiated.  The following information of the hand clusters will be preserved
from the initialization:
\begin{itemize}
\item $ID_{1}$ and $ID_{2}$: Region IDs of both hand clusters,
\item $\p_{1}$ and $\p_{2}$: Homogeneity measure of both hand clusters,
\item $pos_{1}$ and $pos_{2}$: Cluster center of the first and the second hand in
  3D, with regard to the camera's optical center.
\end{itemize}
All these values are required in the tracking phase.

\subsection{Frame-to-Frame Tracking}
\label{s:tracking:track}

\begin{figure}[t]
  \begin{center}
    \begin{tabular}{cc}
      \subfigure[]{
        \includegraphics[width=.4\linewidth]{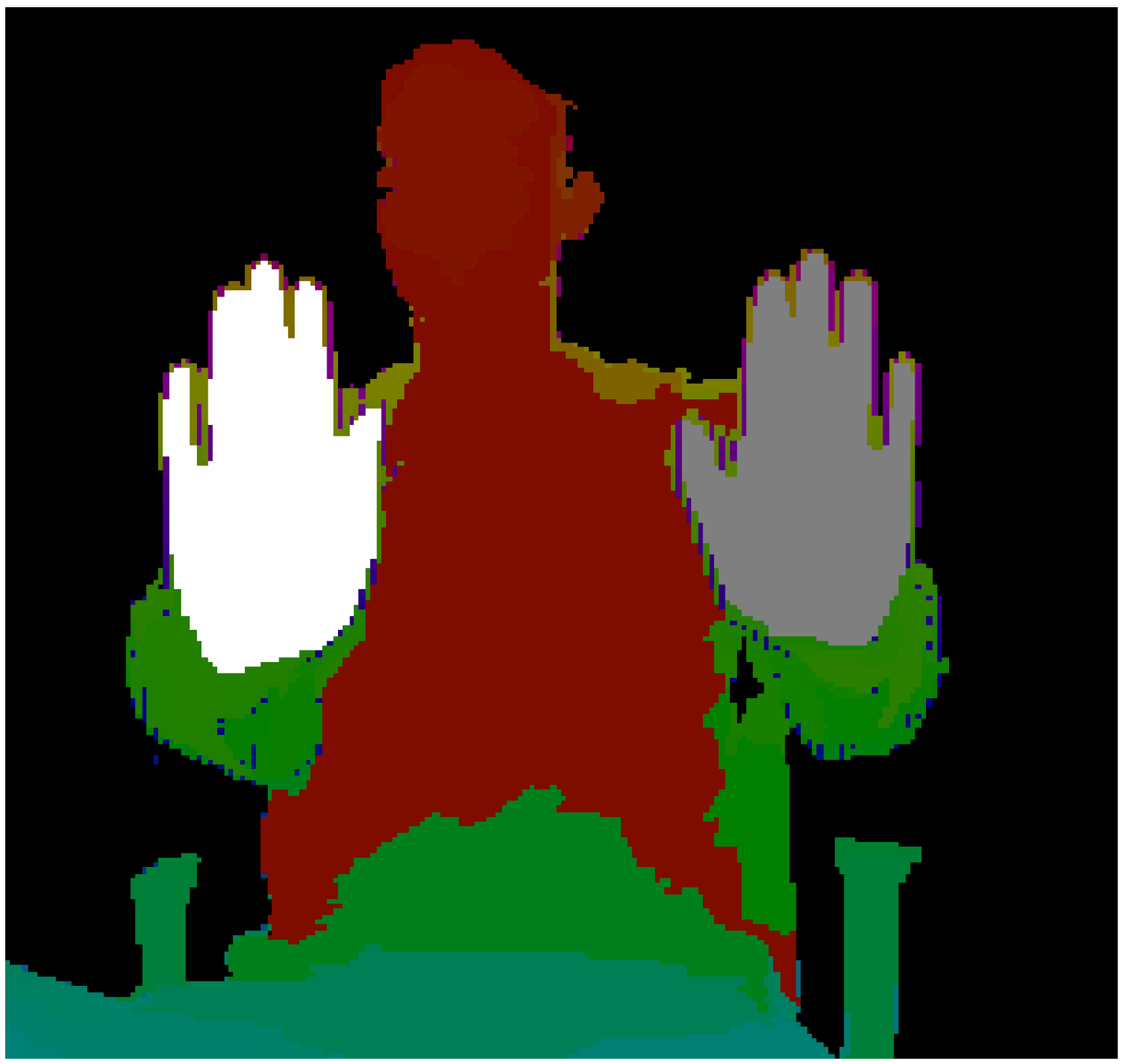}
        \label{fig:rc.1}
      }&
      \subfigure[]{
        \includegraphics[width=.4\linewidth]{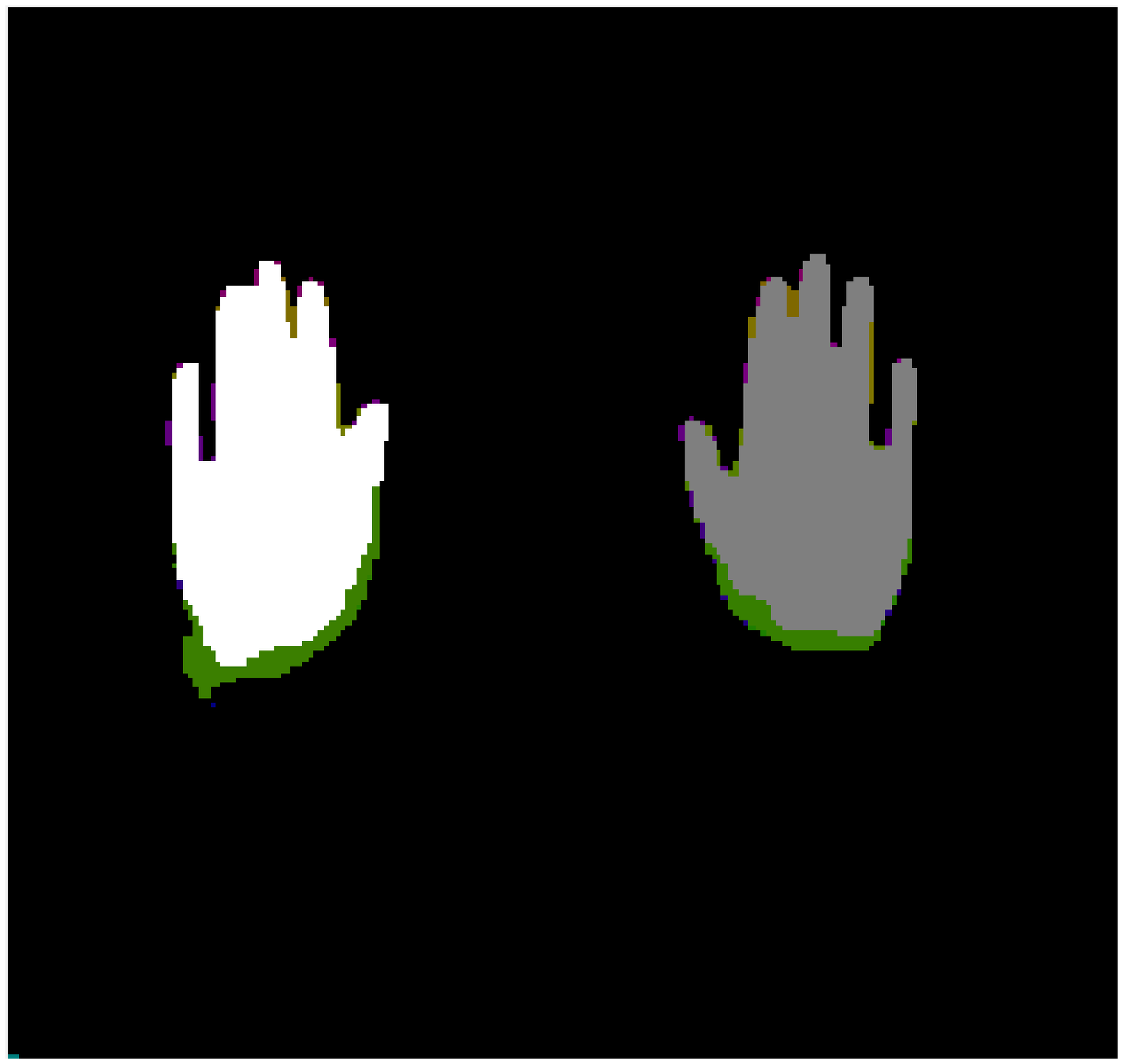}
        \label{fig:rc.2}
      }
    \end{tabular}

  \end{center}
  \caption{Clustered image with range clipping. (a) last frame of the
    initialization. (b) first frame of the tracking step.}
  \label{fig:rc}
\end{figure}

In step two, the maximum $r_{max}$ and minimum $r_{min}$ camera range will be
limited to the hands (see Fig. \ref{fig:rc}). Thus, pixels outside the range
$[r_{min},r_{max}]$ will be ignored for clustering, resulting in a speed up
for the clustering process.  

\begin{figure}[b!]
  \begin{center}
    \subfigure[]{
      \includegraphics[width=.6\linewidth]{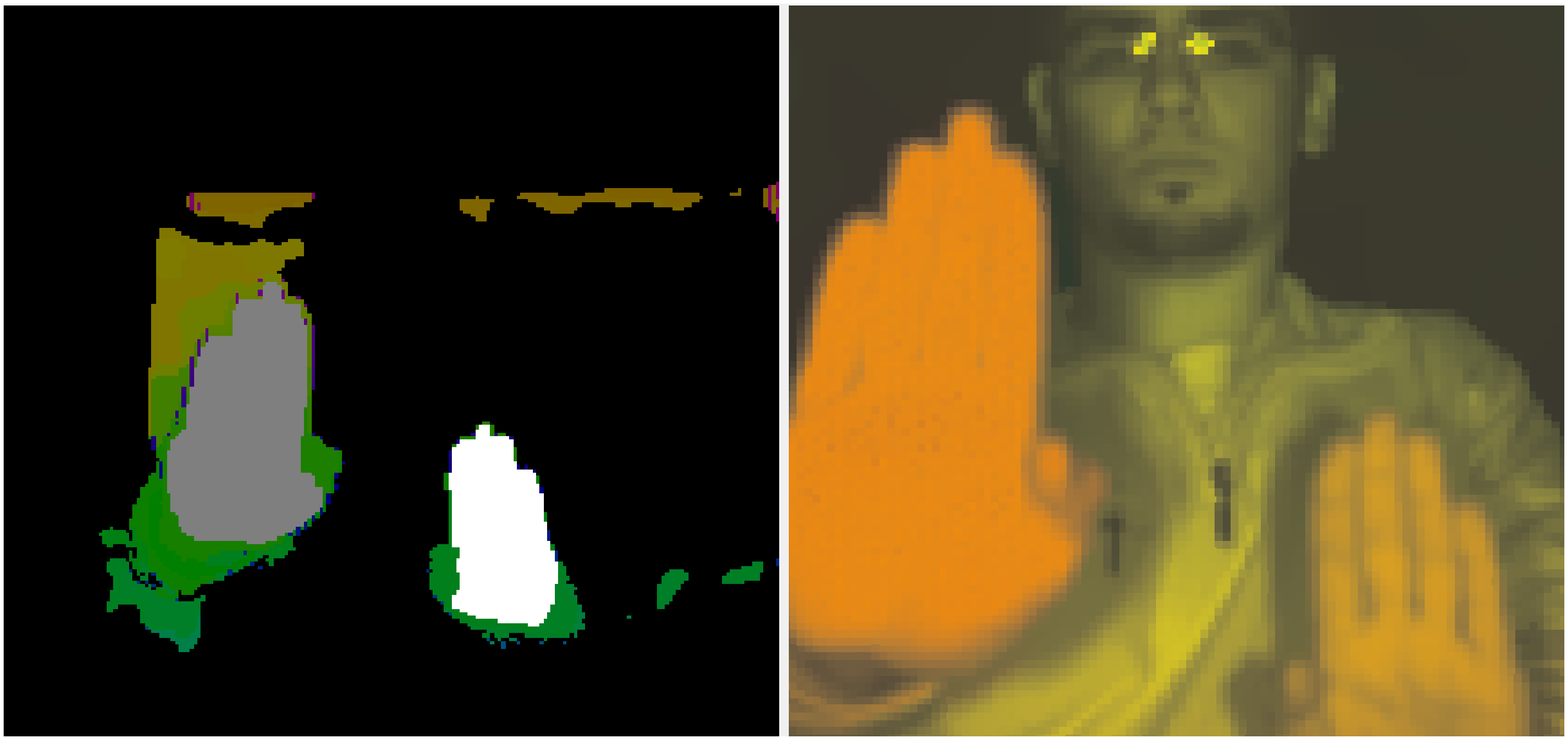}
      \label{fig:backforth.1}
    }
    \subfigure[]{
      \includegraphics[width=.6\linewidth]{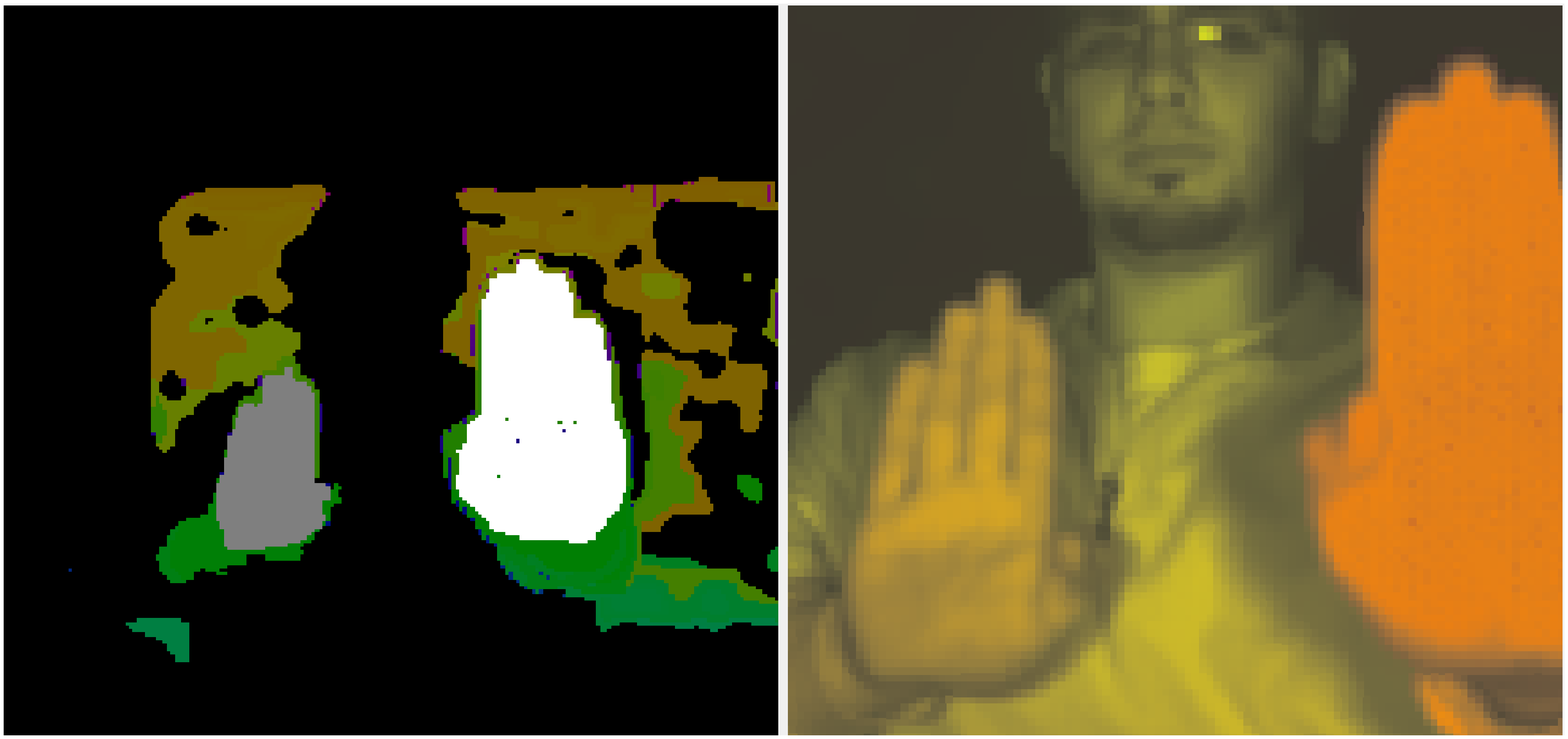}
      \label{fig:backforth.2}
    }
    \caption[Tracking: back and forth]{Hand tracking through nearest neighbor
      assignment. Back and forth movement of the hands.}
    \label{fig:backforth}
  \end{center}
\end{figure}

$r_{min}$ and $r_{max}$ are recalculated for every frame as follows:
\begin{eqnarray*}
  r_{min}&=&min(d_{1},d_{2}) - r_{th}\\
  r_{max}&=&max(d_{1},d_{2}) + r_{th}
\end{eqnarray*}
where $d_{1}, d_{2}$ are the average distances of the first and the second
hand and $r_{th}$ a distance threshold. $r_{th}$ is needed, so that the user
can move his hands freely back and forth and avoids, that the hands could be
clipped away in the next frame (see Fig.~\ref{fig:backforth}).

After setting the minimum and maximum range, two new hand clusters will be
searched in the next frame. All the clusters whose size are above a given minimum size $size_{min}$, will be compared to the hand
clusters from the last frame, based on their homogeneity measure $\p$. From
the set of possible hand clusters, two clusters will be then assigned as the
new hands through a nearest neighbor method with regard to the tracking information
$(ID, \p, pos$) from the hand clusters identified in the last frame.
\subsection{Mutual Occlusion of Hands}
\label{s:tracking:special}

The tracking method described in Sec.~\ref{s:tracking:track} works robustly as
long as two hands are clearly visible and distinguishable in the ToF-images.
Crossing hands would, however, lead to problems, because one hand would be
temporarily covered by the other hand, but the tracking algorithm would still
search for the second hand.

Therefore, to allow that one hand temporary gets covered by the other hand,
the following control method has been added to the tracking algorithm.

Right after the hands have been detected in the initialisation step (see
Sec.~\ref{s:tracking:init}), the algorithm checks the $XY$-distance between
both hands in every frame. Depending on this distance it will be decided if
both hands are in a critical area $D$, where the possibility of covering one
hand with the other hand in one of the subsequent frames is high. $D$ is defined
as follows:
\[
D=\begin{cases}
  true,  & \Betr{(pos_{1})_{xy} - (pos_{2})_{xy}} < d_{min} \\
  false, & \text{else.}
\end{cases} 
\]
where $(pos_{i})_{xy}$ is the $XY$-position of the $i$-th hand and $d_{min}$
is the minimal distance.

As long as the distance between both hands is above the threshold $d_{min}$, the
hands will be tracked as described in Sec.~\ref{s:tracking:track}). In case if
$ D=true$, the following procedure is initiated:

\paragraph{Check for possible disappearance of a single hand}
A hand is considered missing, if in the current frame:
\begin{itemize}
\item no cluster has been found which can resemble the hand (based on $\p$),
  or
\item a cluster has been found, but the distance between its cluster center
  and the hand position of the last frame, is above a given maximal distance,
  implying that no cluster can be assigned to the hand in a meaningful way.
\end{itemize}

\paragraph{Detection and tracking of the uncovered hand}
After a hand gets covered by the other, the last resolved tracking information
of the disappeared hand cluster (backhand cluster $\cR_{bh}$) is stored in
order to detect the lost hand again after reappearance. The front hand cluster
$\cR_{fh}$, however, will be continuously tracked as described in
Sec.~\ref{s:tracking:track}.  The lost hand will be searched in the set of
remaining clusters. In order to reassign the lost hand to a cluster $\cR$ the
following criteria have to be satisfied:
\begin{itemize}
\item The cluster center must be behind the front hand cluster center, in
  $Z$-direction, i.e. $(pos_{\cR})_z < ({pos_{\cR}}_{fh})_z$.
\item The distance between the cluster center $pos_{\cR}$ and the center of the
  front hand cluster ${pos_{\cR}}_{fh}$, must not exceed a threshold $t_d$: $\left|
    pos_{\cR} - {pos_{\cR}}_{fh}\right| < t_d$

  Thus the lost hand appears beside the hiding hand.
\item The difference between the homogeneity measures of the cluster $\cR$
  and the stored measure of the back hand cluster $\cR_{bh}$, must not exceed a
  threshold $t_{\p}$: $\Betr{{\p_\cR}_{bh}-{\p_\cR}} < t_{\p}$.
\end{itemize}

If more than one cluster satisfy these criteria, the new uncovered hand cluster
will be assigned to the one with minimal $\p$ deviation. If none of the
clusters satisfy these criteria, the search for the lost hand will continue in
the next frame.


\section{Results}
\label{s:results}

\begin{table}
  \begin{center}
    \begin{tabular}{|l|c|c|}
      \hline
      Parameters & Value \\
      \hline\hline
      $t_z$    	& $0.04m$\\
      $t_{\p}$  & $0.009$ (rad)\\   
      $\alpha_z$    	& $\frac{8}{\pi}$\\
      $\alpha_{\p}$  & $\frac{4}{3}$\\      
      \hline\hline
      $size_{min}$    & $200px$\\
      $t_d$        		& $0.1m$\\
      $t_{\p}$        & $0.009$ (rad)\\
      $r_{th}$				& $0.1m$\\
      $d_{min}$      	& $0.1m$\\
      \hline
    \end{tabular}
  \end{center}
  \caption{Parameters used for the clustering (first four rows) and hand tracking algorithm (rest).} 
  \label{tab:ths}
\end{table}

\begin{table}
  \begin{center}
    \begin{tabular}{|l|c|c|}
      \hline
      Process & time (Init) & time (Tracking) \\
      \hline\hline
      Update Camera         & 16 & 16 \\
      \hline\hline
      Find Mergepartner     & 29 & 27\\
      Merge Regions         & 5  & 3\\
      Update Values         & 10  & 4\\
      \hline\hline
      Tracking   			 	    & 7  & 9\\
      \hline\hline
      Total Time            & 67 & 59\\
      \hline
    \end{tabular}
  \end{center}
  \caption{Average runtime (in msec) on an Intel Core 2 Duo PC with
    3.00 GHz, 4 GB RAM and a NVIDIA GeForce GTX 480 graphic card, for 
    $204^2px$ ToF-data.} 
  \label{tab:speed}
\end{table}

\begin{figure}[t]
  \begin{center}
    \begin{tabular}{cc}
      \subfigure[]{
        \includegraphics[width=.4\linewidth]{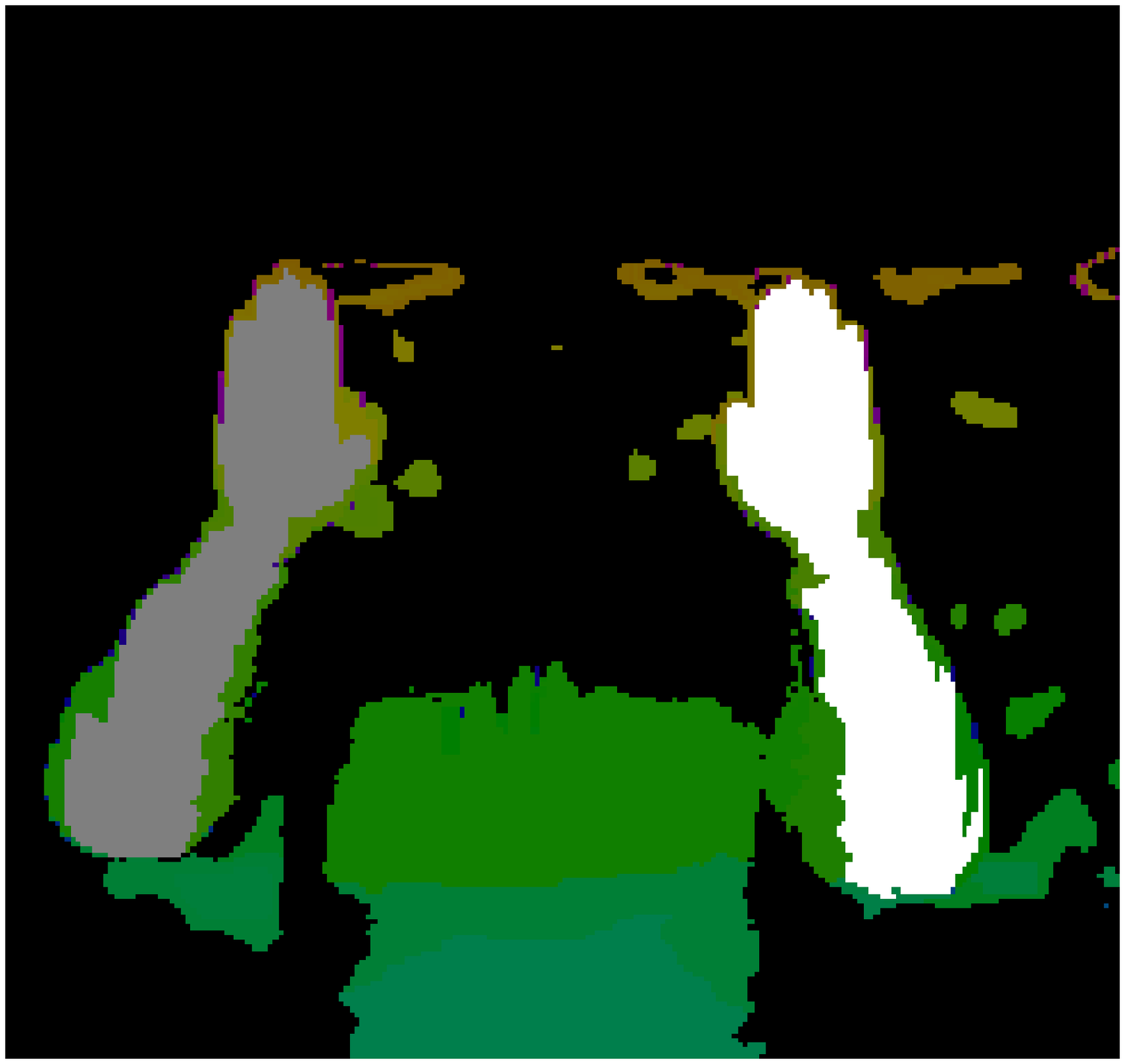}
        \label{fig:cerr.1}
      }&
      \subfigure[]{
        \includegraphics[width=.4\linewidth]{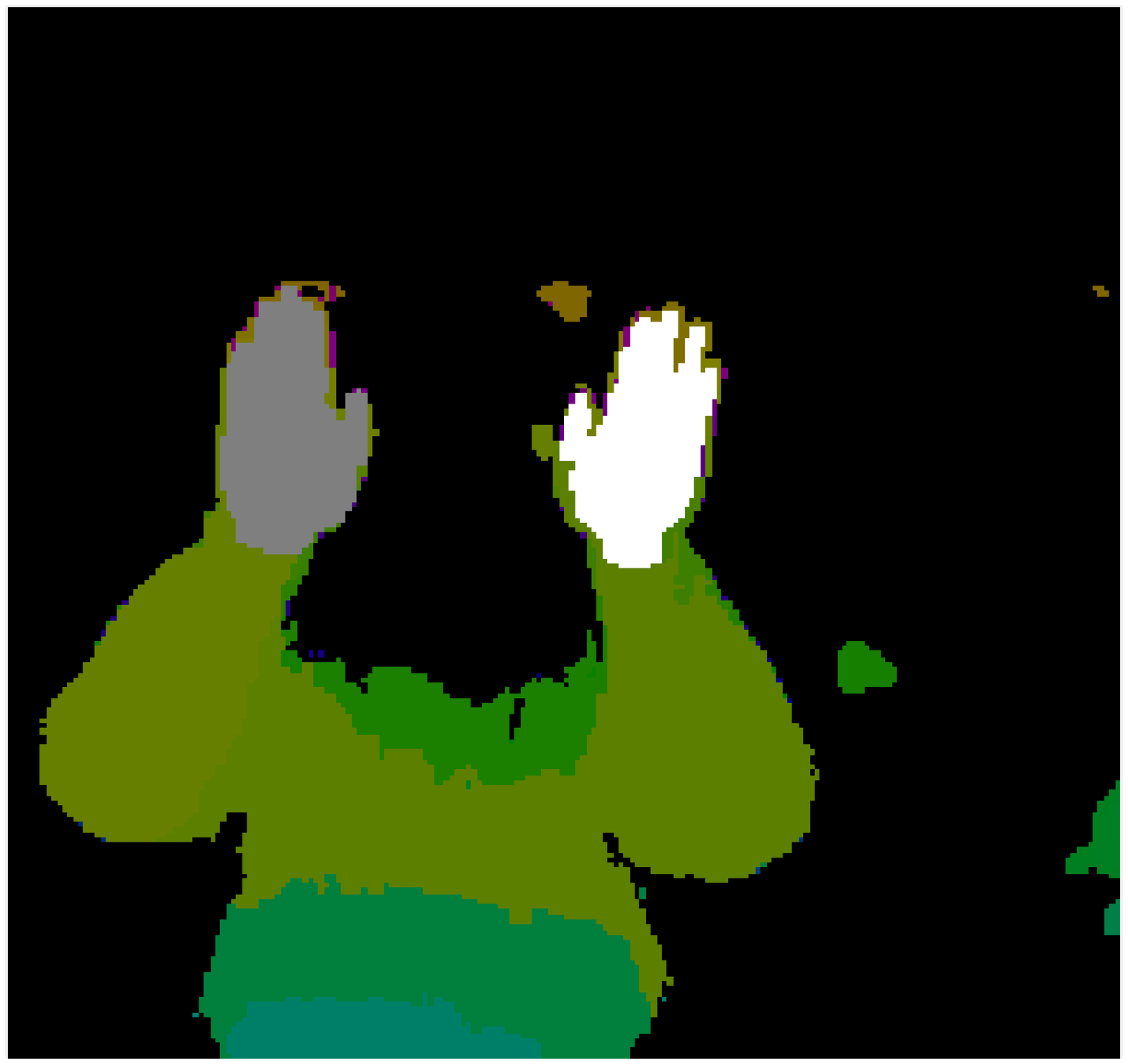}
        \label{fig:cerr.2}
      }
    \end{tabular}
  \end{center}
  \caption{User holds both hands up. Hand palms and arms are on the same
    $XY$-plane. The grey and the white clusters are the hand clusters. (a)
    user wears a skin-colored sweater. The arms and hands get clustered
    together. (b) user wears a brown leather jacket. Arms and hands are
    clustered separately.}
  \label{fig:cerr}
\end{figure}

After both hands have been detected in the initialization, the clustering and
tracking algorithm runs with an average frame-rate of 16 frames/sec
(see Table \ref{tab:speed}). However, some restrictions must be satisfied, to
ensure a flawless hand tracking. Objects or bodyparts, with a similar
reflectivity as skin must not get in contact with a hand, i.e. touch the hand
in approximately the same $Z$-distance to the ToF-camera, otherwise they will
be assigned to one hand cluster (see Fig.~\ref{fig:cerr}).

\begin{figure}[t]
  \begin{center}
    \begin{tabular}{ccc}
      \subfigure[]{
        \includegraphics[width=.29\linewidth]{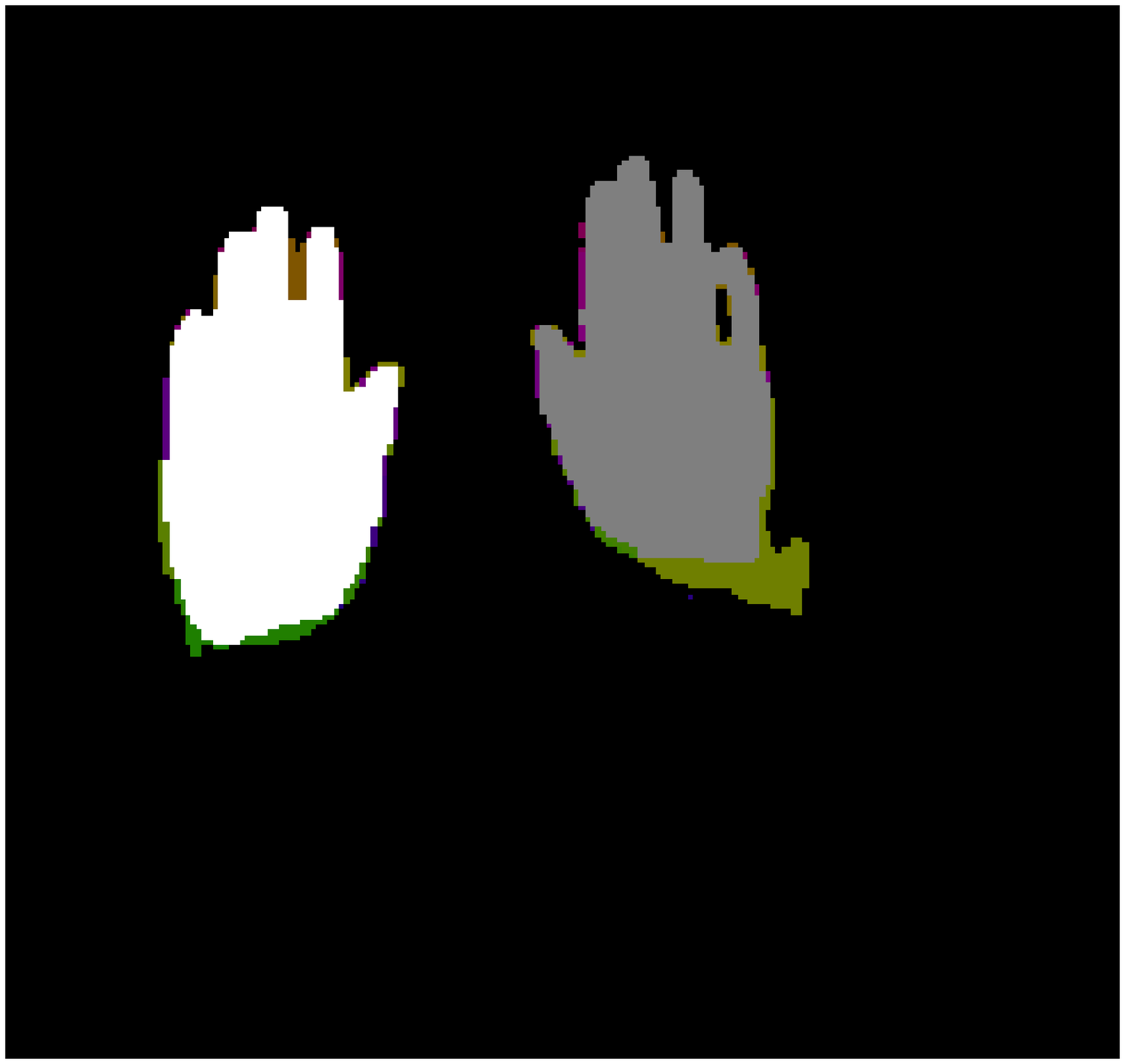}
        \label{fig:terr.1}
      }&
      \subfigure[]{
        \includegraphics[width=.29\linewidth]{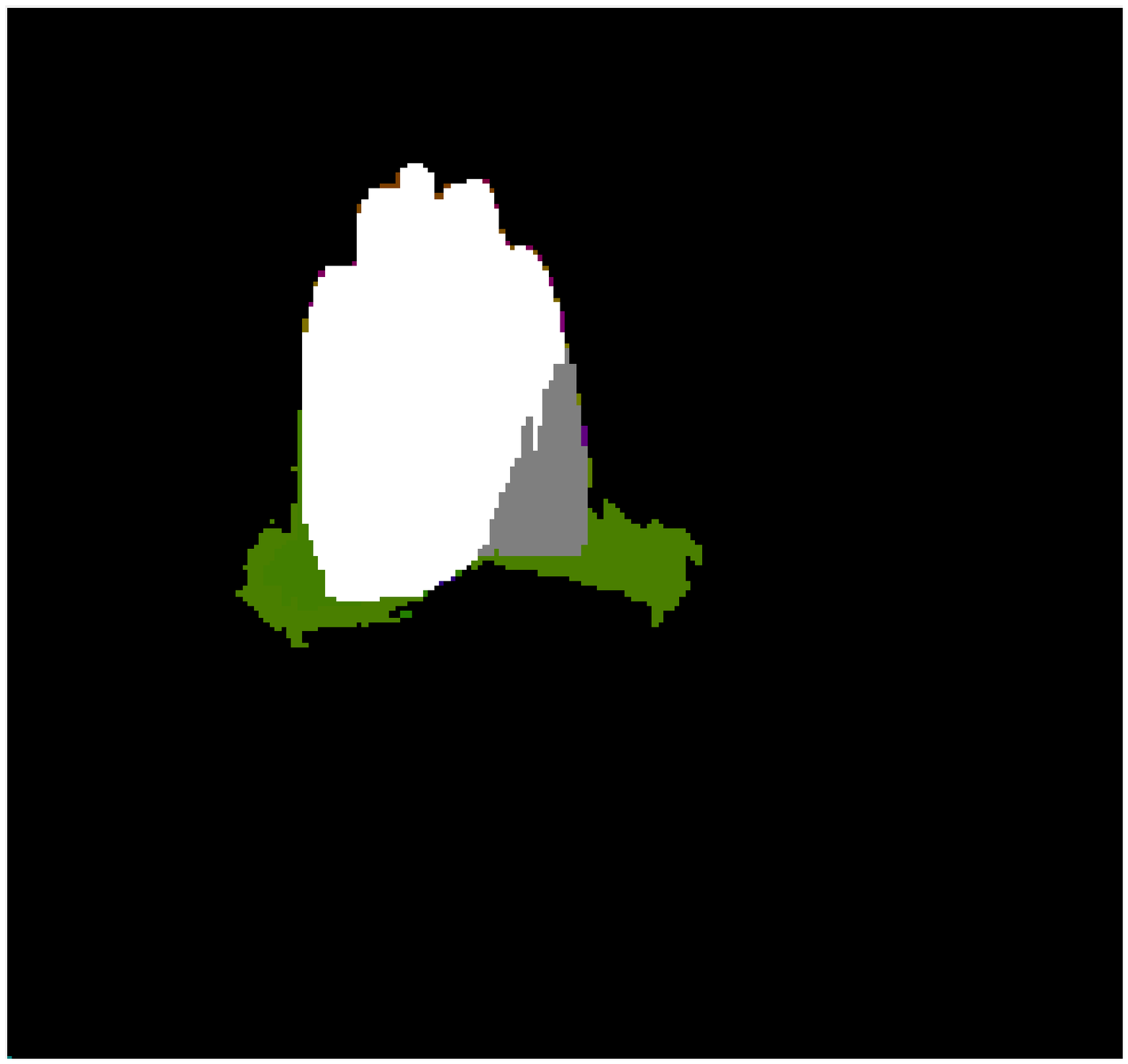}
        \label{fig:terr.2}
      }&
      \subfigure[]{
        \includegraphics[width=.29\linewidth]{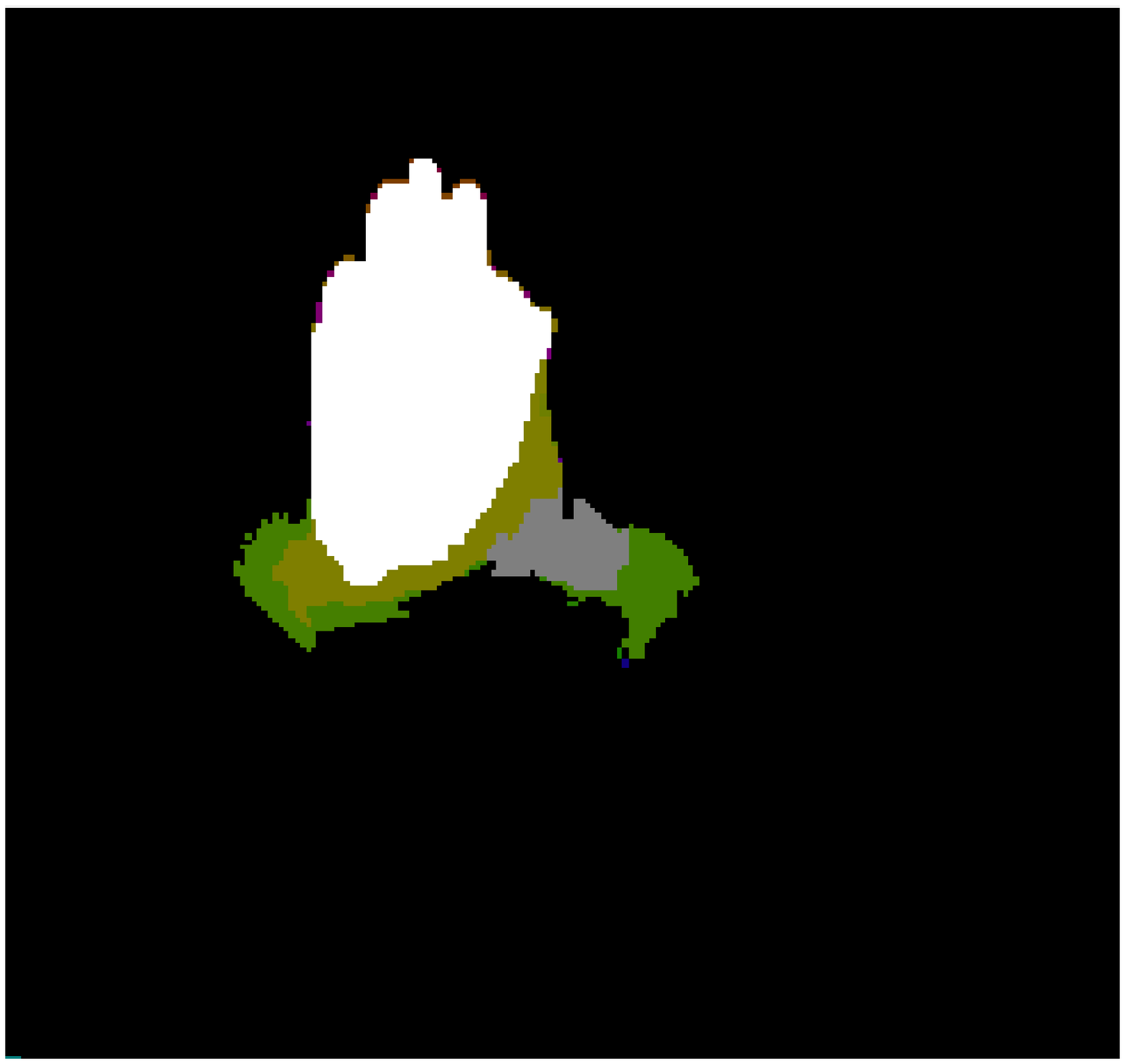}
        \label{fig:terr.3}
      }
    \end{tabular}
  \end{center}
  \caption{(a)-(c) shows a crossing sequence of both hands, where the hand are
    close together also in $Z$-direction. (b) Since both hands get in contact
    while crossing each other, a part of the back hand gets clustered to the
    front hand cluster, leading to tracking errors (as seen in (c)). (c) because the center of the back hand was shifted, the wrist gets mistaken as the new hand cluster.}
  \label{fig:terr}
\end{figure}

A similar error occurs when crossing both hands. Being the fact, that both
hands usually have the exact same homogeneity measure $\p$, not keeping a
given minimum distance in $Z$-direction while crossing, can lead to a merging
of both hands into a single hand cluster (see Fig.~\ref{fig:terr}). These
clustering issues can forthermore lead to tracking errors
(see Fig. \ref{fig:terr.3}).

\begin{figure}[t]
  \begin{center}
      \subfigure[]{
        \includegraphics[width=.6\linewidth]{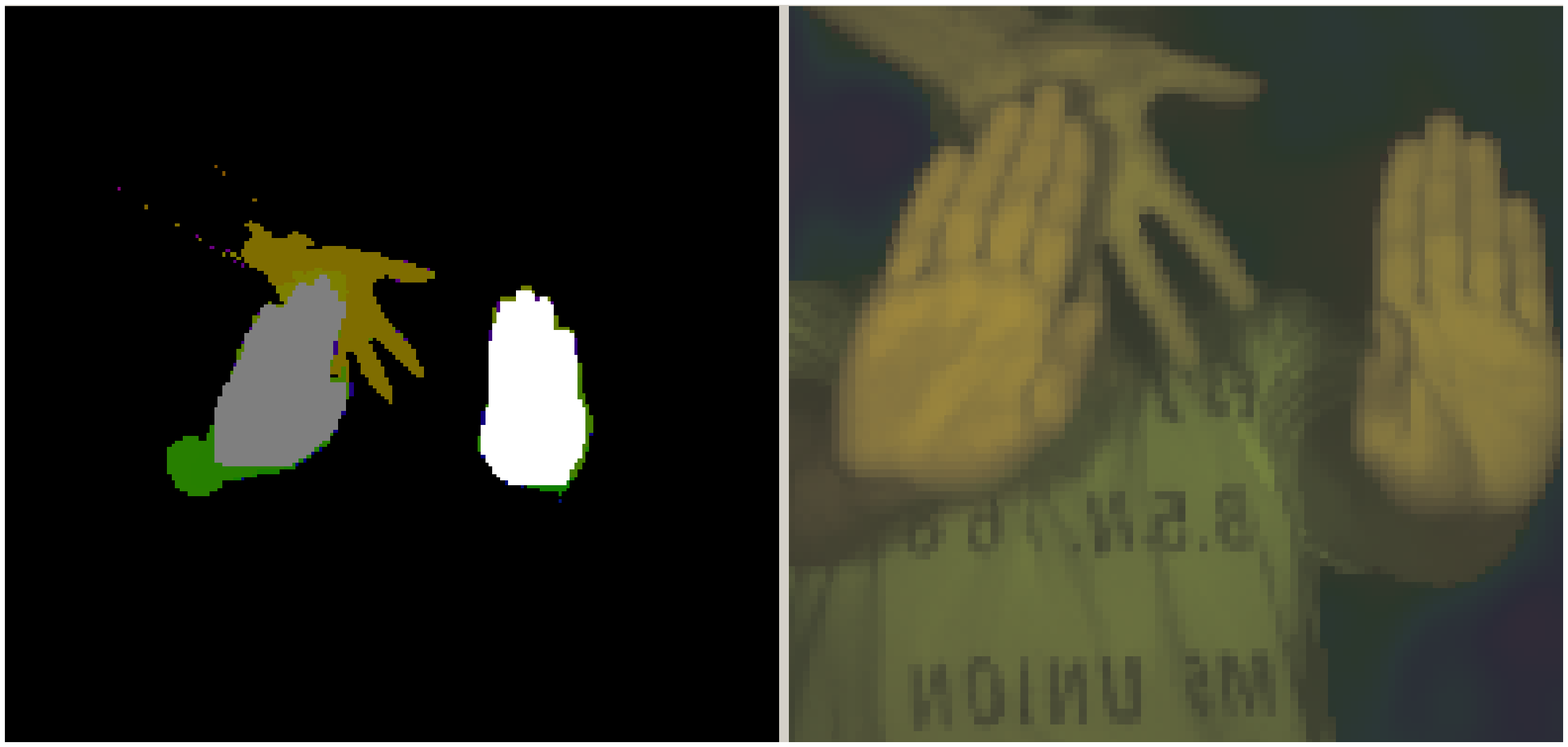}
        \label{fig:3rdh.1}
      }
      \subfigure[]{
        \includegraphics[width=.6\linewidth]{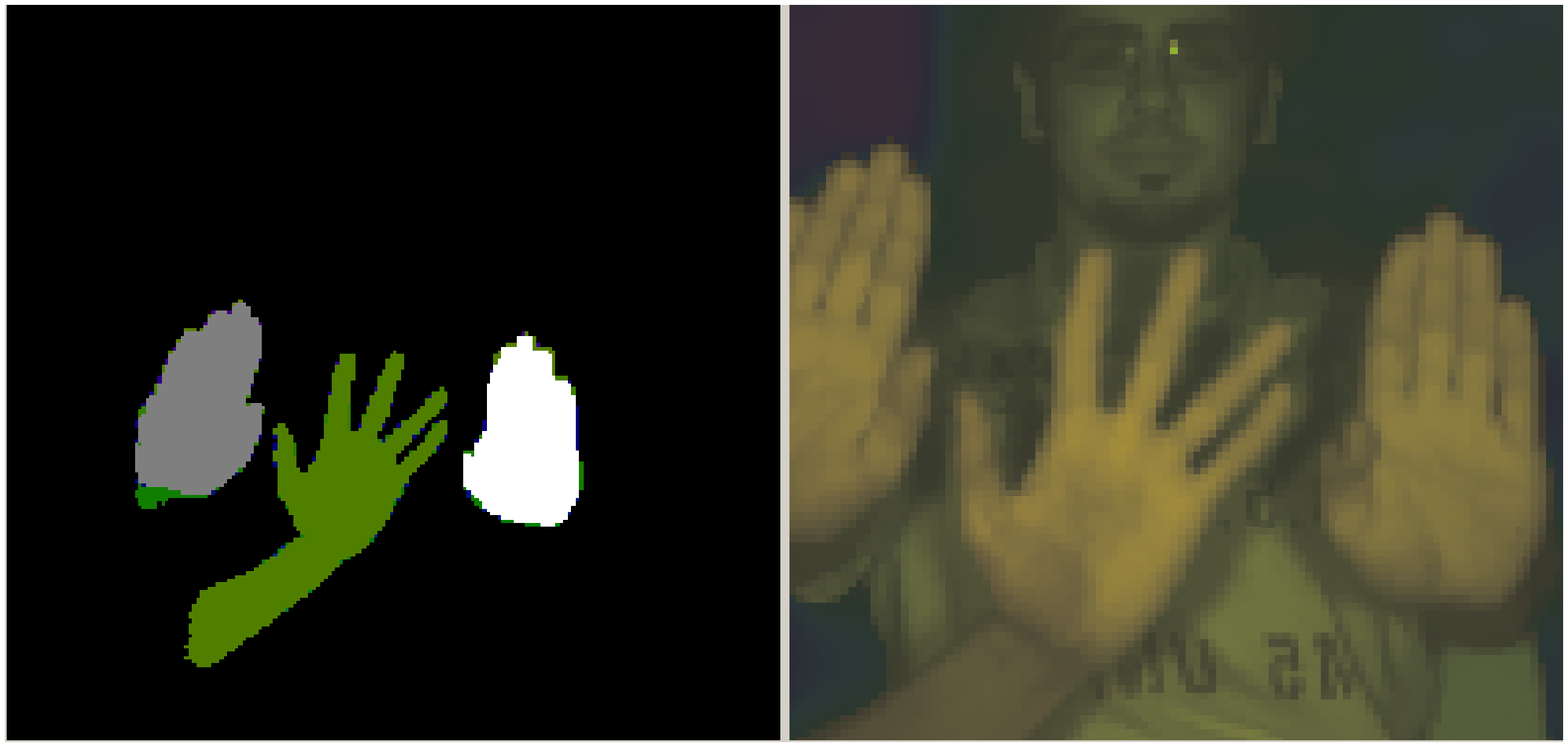}
        \label{fig:3rdh.2}
      }
      \subfigure[]{
        \includegraphics[width=.6\linewidth]{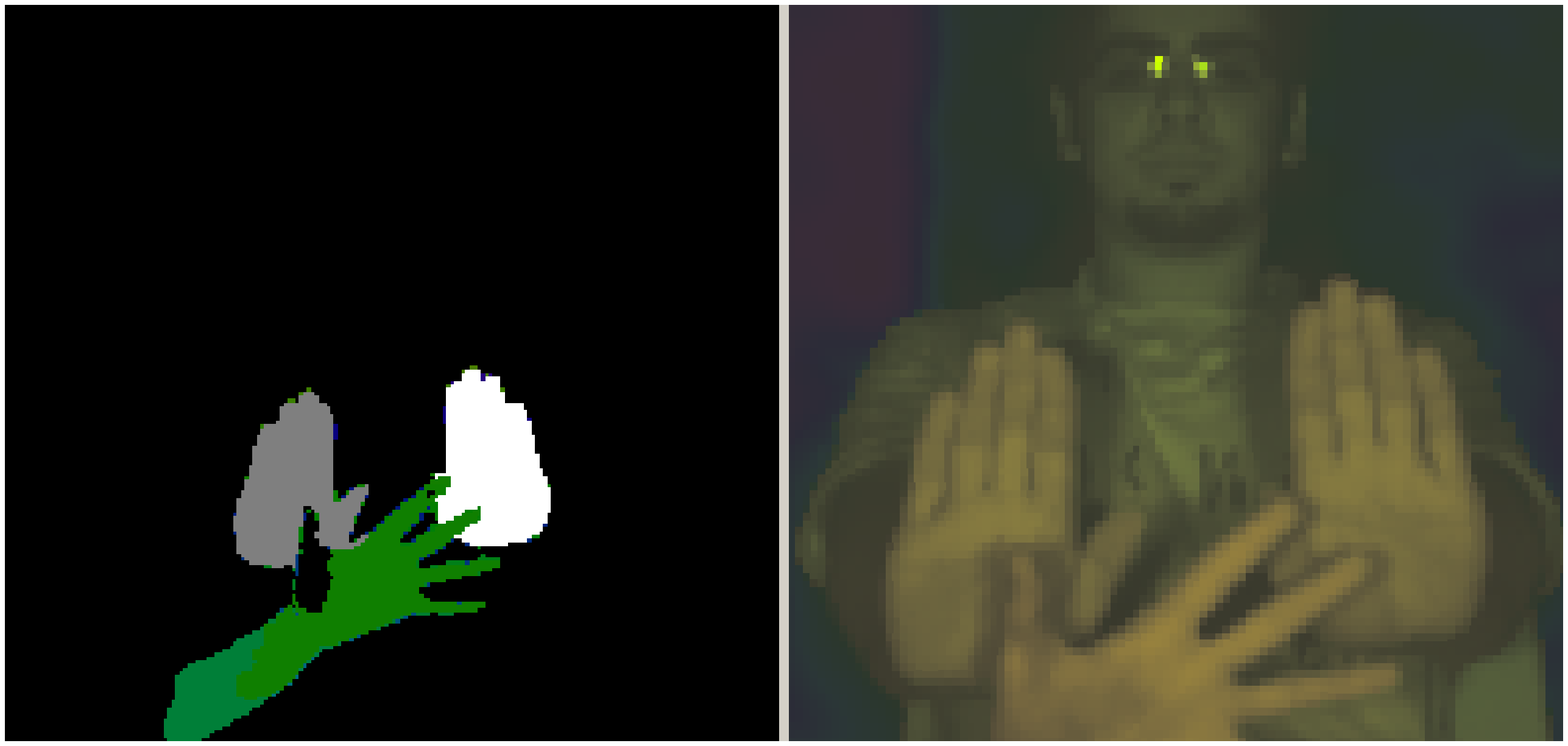}
        \label{fig:3rdh.3}
      }
  \end{center}
  \caption{The user's visible hands will be tracked correctly, although a third hand is behind (a), in between (b), or in front (c) of the user's tracked hands.}
  \label{fig:3rdh}
\end{figure} 

However, as long as the mentioned restrictions are satisfied, the tracking algorithm can handle a third hand and will always track the correct two visible hands (see Fig. \ref{fig:3rdh}). 

\begin{figure}[t]
  \begin{center}
    \begin{tabular}{cc}
      \subfigure[]{
        \includegraphics[width=.4\linewidth]{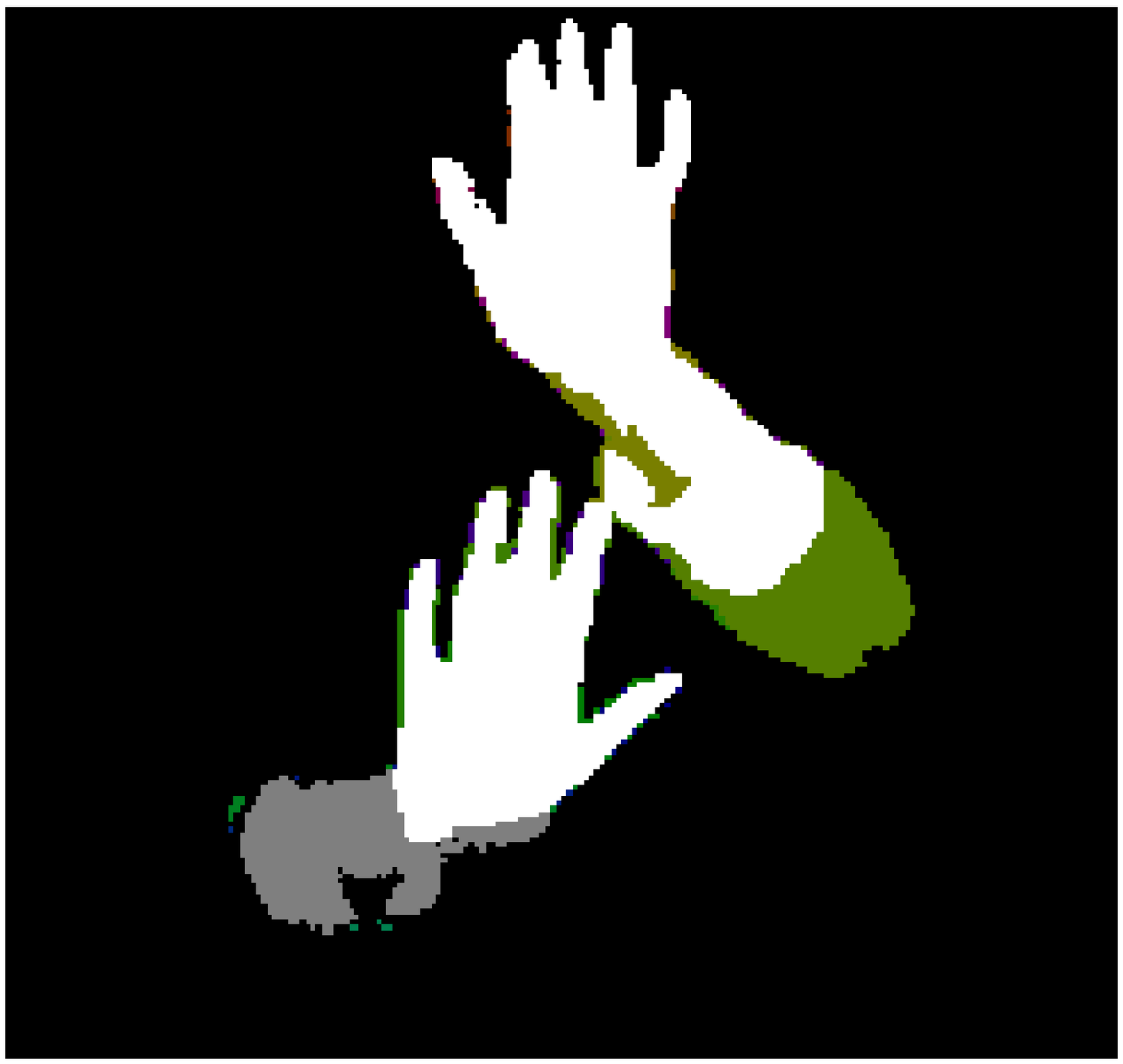}
        \label{fig:phi.1}
      }&
      \subfigure[]{
        \includegraphics[width=.4\linewidth]{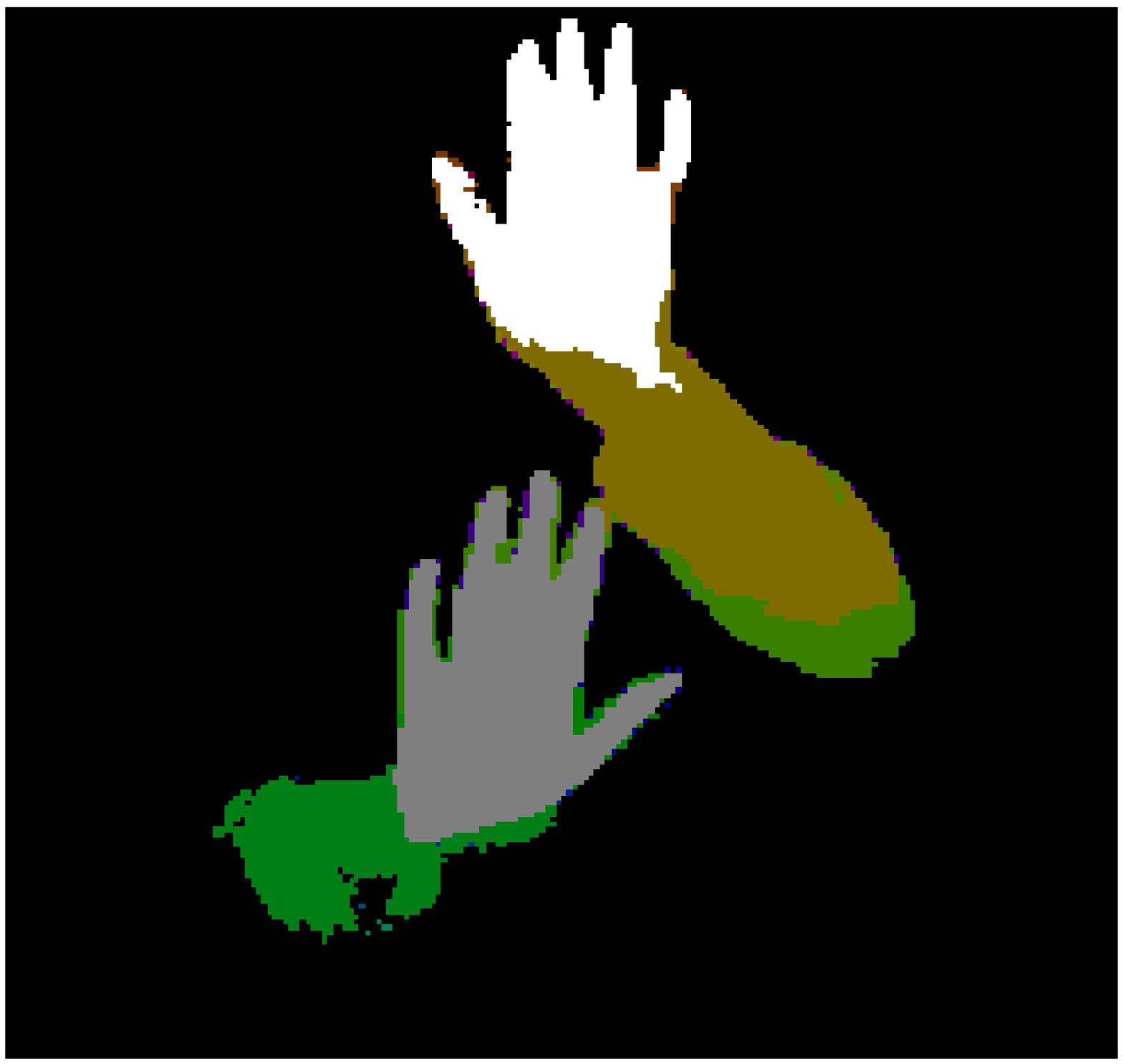}
        \label{fig:phi.2}
      }
    \end{tabular}
  \end{center}
  \caption{(a) clustering result with $\p$ from Ghobadi \etal
    Eq.~\eqref{eq:ZessPhi}. Both hands get clustered into one hand cluster
    (white cluster) by a connection via the sleeve. The other sleeve gets then
    mistaken as the second hand cluster (grey cluster). (c) clustering result
    with $\p$ from Eq.~\eqref{eq:phi} as proposed in this paper. Both hands
    get clustered correctly.}
  \label{fig:phi}
\end{figure}

As mentioned in Sec.~\ref{s:clustering:mc}, the homogeneity measure $\p$ is
based on the idea of Ghobadi~\etal~\cite{GLHL07:HS}. The main difference
between both measures is, that Ghobadi~\etal chose a linear approach to
describe the relation between distance and intensity of a given cluster $\cR$:
\begin{equation}
  \label{eq:ZessPhi}
  \p_{\cR} = \arctan{\left(\frac{d_{\cR}}{I_{\cR}}\right)},
\end{equation} 
where $d_{\cR}$ is the normalized distance and $I_{\cR}$ the normalized
intensity of the cluster. This approach can lead to inacurate clustering
results (see Fig.~\ref{fig:phi}). With a more accurated calculation of $\p$
from Eq.~\eqref{eq:phi} those clustering errors are significantly reduced.\\

A list with all the parameters and their values, which were applied, are shown in Table \ref{tab:ths}.

\noindent

\section{Conclusion}
\label{s:conclusion}

In this paper, we introduce a fast an robust hand segmentation approach based
on a hierarchical clustering technique, which is implemented on the GPU as a
parallel merging algorithm in order to achieve a high performance.  We also
introduce a novel homogeneity criterion robustly merging regions in fused
range-intensity images. The intensity and range information is fused into one
pixel value, representing its combined intensity-depth homogeneity.

The results show that when certain restrictions are satisfied, our hand
segmentation and tracking system is capable of robustly detecting and tracking
both hands in real-time, even under the condition that one hand is temporarly
covered by the other hand.


\newcommand{\ACG}{Advanced Computer Graphics}
\newcommand{\AM}{Aequationes Math.}
\newcommand{\ASIM}{Proc. Symposium on Simulation Technique}
\newcommand{\BAMS}{Bull. Amer. Math. Soc.}
\newcommand{\BIT}{BIT}
\newcommand{\C}{Computing}
\newcommand{\CA}{Constr. Approx.}
\newcommand{\CAD}{Computer Aided Design}
\newcommand{\CADCG}{CAD und Computergraphik}
\newcommand{\CAGD}{Computer-Aided Geom. Design}
\newcommand{\CGF}{J. Computer Graphics Forum}
\newcommand{\CGTA}{Computational Geometry, Theory and Applications}
\newcommand{\CJ}{Computer J.}
\newcommand{\CGIP}{Computer Graphics and Image Processing}
\newcommand{\CVGIP}{Computer Vision Graphics and Image Processing}
\newcommand{\CVPR}{IEEE Conf. on Computer Vision and Pattern Recognition}
\newcommand{\CVUI}{J. Computer Vision and Image Understanding}
\newcommand{\EG}{J. Computer Graphics Forum (Proc. Eurographics)}
\newcommand{\EGSHORT}{Proc. Eurographics, Short-Paper}
\newcommand{\EGRW}{Eurographics Workshop on Rendering}
\newcommand{\EGSR}{Eurographics Symposium on Rendering}
\newcommand{\GI}{Proc. Graphics Interface}
\newcommand{\GIIT}{Proc. GI-Informatiktage}
\newcommand{\HWWS}{Proc. ACM/Eurographics Graphics Hardware}
\newcommand{\HPG}{Proc. ACM/Eurographics High Performace Graphics}
\newcommand{\ICGA}{IEEE Computer Graphics \& Appl.}
\newcommand{\IEEE}{IEEE Computer Graphics \& Appl.}
\newcommand{\IMAN}{Inst. Math. Applics. Numer. Anal.}
\newcommand{\IND}{Ind. Univ. J. Math.}
\newcommand{\IGARSS}{IEEE Int. Geosc. \& Remote Sensing Symp. (IGARSS)}
\newcommand{\IRS}{Conf. Proc. Intern. Radar Symposium}
\newcommand{\IVC}{J. Image and Vision Computing}
\newcommand{\JACM}{J. ACM}
\newcommand{\JAM}{J. Analyses Math.}
\newcommand{\JAT}{J. Approx. Th.}
\newcommand{\JCAM}{J. Comp. Appl. Math.}
\newcommand{\JCV}{J. of Computer Vision}
\newcommand{\JIMA}{J. Inst. Math. Applics.}
\newcommand{\JMA}{SIAM J. Math. Anal.}
\newcommand{\JMAA}{J. Math. Anal. Appl.}
\newcommand{\JMM}{J. Math. Mech}
\newcommand{\JMP}{J. Math. Phys.}
\newcommand{\JSSC}{J. Sci. Stat. Comp.}
\newcommand{\LAA}{Linear Alg. Appl.}
\newcommand{\MAA}{Math. Anal. Appl.}
\newcommand{\MC}{Math. Comp.}
\newcommand{\MMNA}{Mathematical Modelling and Numerical Analysis}
\newcommand{\NM}{Numer. Math.}
\newcommand{\NFAO}{Numer. Func. Anal. Optim.}
\newcommand{\PAMI}{IEEE Trans. Pattern Anal. and Mach. Intell.}
\newcommand{\PAMS}{Proc. Amer. Math. Soc.}
\newcommand{\PEMS}{Proc. Edinburgh Math. Soc.}
\newcommand{\PG}{Proc. Pacific Graphics}
\newcommand{\PJM}{Pacific J. Math.}
\newcommand{\RMJ}{Rocky Mt. J. Math.}
\newcommand{\SIG}{ACM Trans. Graph. (Proc. SIGGRAPH)}
\newcommand{\SIMPRA}{J. Simulation Practice \& Theory}
\newcommand{\SJNA}{SIAM J. Numer. Anal.}
\newcommand{\SMA}{SMA}
\newcommand{\SPIE}{Proc. SPIE}
\newcommand{\SPIEVIP}{Proc. SPIE Visual Information Processing}
\newcommand{\TAMS}{Trans. Amer. Math. Soc.}
\newcommand{\TCVG}{EG/IEEE Symopsium on Visualization}
\newcommand{\TOG}{ACM Trans. Graph.}
\newcommand{\TGRS}{IEEE Transactions on Geoscience and Remote Sensing}
\newcommand{\TOMS}{ACM Trans. Math. Software}
\newcommand{\TVCG}{IEEE Trans. on Visualization and Computer Graphics}
\newcommand{\VC}{The Visual Computer}
\newcommand{\VMV}{Proc. Vision, Modeling and Visualization}
\newcommand{\VIS}{Proc. IEEE Conf. on Visualization}
\newcommand{\WSCG}{Proc. Conf. on  Computer Graphics, Visualization and
Computer Vision (WSCG)}
\newcommand{\ZAMM}{Zeitschrift f\"ur Angewandte Mathematik und Mechanik}

\newcommand{\SLN}{Lecture Notes in Mathematics}


{\small
  \bibliographystyle{ieee}
  \bibliography{egbib}
}

\end{document}